\documentclass[letterpaper, 10 pt, conference]{ieeetran}  

\IEEEoverridecommandlockouts    

\usepackage{fancyhdr}
\fancypagestyle{firstpage}{%
  \lhead{\footnotesize 2377-3766 (c) 2021 IEEE. Personal use is permitted, but republication/redistribution requires IEEE permission. See \url{http://www.ieee.org/publications_standards/publications/rights/index.html} for more information. DOI: 10.1109/LRA.2022.3141194}
}

\usepackage{graphics} 
\usepackage{amssymb}  
\usepackage{textglos}
\usepackage[utf8]{inputenc} 
\usepackage[T1]{fontenc}    
\usepackage{hyperref}       
\usepackage{url}            
\PassOptionsToPackage{hyphens}{url}\usepackage{hyperref}
\usepackage[hyphenbreaks]{breakurl}
\usepackage{booktabs}       
\usepackage{amsfonts}       
\usepackage{nicefrac}       
\usepackage{microtype}      
\usepackage{lipsum}
\usepackage{tabularx} 
\usepackage{amsmath}  
\usepackage{graphicx,dblfloatfix} 
\usepackage[margin=1in,letterpaper]{geometry} 
\usepackage{cite} 
\usepackage{titlesec}
\usepackage{mathtools, amssymb, nccmath}
\usepackage{wasysym}
\usepackage{algorithm}
\usepackage{algorithmic}
\usepackage{multicol}
\let\labelindent\undefined
\usepackage{enumitem}
\usepackage{float}
\usepackage{mathtools}
\usepackage{amsmath,amsfonts,amssymb}
\usepackage{cmll}
\usepackage{tensor}
\usepackage{booktabs}
\usepackage{tabularx}
\usepackage{makecell}

\usepackage{tablefootnote}
\usepackage{threeparttable}
\usepackage{multirow}

\usepackage[font=small,skip=5pt]{caption}

\usepackage[dvipsnames]{xcolor}

\newcolumntype{Z}{ >{\centering\arraybackslash}X }

\newcommand{\crff}{\,\overline{\!\times\!}{}^{\,*}}

\newcommand{\M}{\mathcal{M}}

\newcommand{\I}{\mathcal{I}}

\newcommand{\T}{^\top}

\newcommand{\R}{\mathbb{R}}
\newcommand{\C}{\vC}

\newcommand{\rmvec}[1]{\boldsymbol{#1}}
\newcommand{\greekvec}[1]{\boldsymbol{#1}}

\newcommand{\B}{\boldsymbol{B}{}}
\renewcommand{\C}{\boldsymbol{C}}
\renewcommand{\M}{\boldsymbol{M}}

\renewcommand{\I}{\boldsymbol{I}{}}
\newcommand{\g}{\rmvec{g}}
\newcommand{\q}{\rmvec{q}}
\renewcommand{\b}{\rmvec{b}}
\renewcommand{\u}{\rmvec{u}}

\newcommand{\m}{\rmvec{m}}
\newcommand{\f}{\rmvec{f}}
\newcommand{\p}{\rmvec{p}}

\renewcommand{\v}{\rmvec{v}}
\newcommand{\vJ}[1]{\v_{J_{#1}}}

\renewcommand{\t}{\rmvec{t}}

\renewcommand{\a}{\rmvec{a}}

\newcommand{\qd}{\dot{\q}}
\newcommand{\qdd}{\ddot{\q}}

\newcommand{\Psibar}{\greekvec{\Psi}}
\newcommand{\Psibardot}{\,\dot{\!\Psibar}}
\newcommand{\Psibarddot}{\,\ddot{\!\Psibar}}

\newcommand{\taubar}{\greekvec{\tau}}
\newcommand{\gammabar}{\greekvec{\gamma}}

\newcommand{\etabar}{\greekvec{\eta}}
\newcommand{\xibar}{\greekvec{\xi}}
\newcommand{\zetabar}{\greekvec{\zeta}}

\newcommand{\red}[1]{{\color{black}#1}}
\newcommand{\blue}[1]{{\color{black}#1}}

\newcommand{\phibar}{\boldsymbol{s}}
\newcommand{\Phibar}{\boldsymbol{S}}

\newcommand{\Phibardot}{\,\dot{\!\Phibar}}

\def\autodiff{AD}
\newcommand{\subtree}{\nu}
\newcommand{\subtreeb}{\overline{\nu}}

\title{\LARGE \bf
Efficient Analytical Derivatives of Rigid-Body Dynamics using Spatial Vector Algebra }

\author{ Shubham Singh$^{1}$, Ryan P. Russell$^{2}$ and Patrick M.~Wensing$^{3}$
\thanks{$^{1}$Graduate Research Assistant, Aerospace Engineering, The University of Texas at Austin, TX-78751, USA. \href{mailto:singh281@utexas.edu}{singh281@utexas.edu}}
       
     \thanks{$^{2}$Associate Professor, Aerospace Engineering, The University of Texas at Austin, TX-78751, USA. \href{mailto:ryan.russell@utexas.edu}{ryan.russell@utexas.edu} 
           }%
       \thanks{$^{3}$Assistant Professor, Aerospace \& Mechanical Engineering, University of Notre Dame, IN-46556, USA. \href{mailto:pwensing@nd.edu}{pwensing@nd.edu}
       }%
    
}

\begin{document}

\maketitle

\thispagestyle{firstpage}
\pagestyle{plain}

\begin{abstract}
An essential need for many model-based robot control algorithms is the ability to quickly and accurately compute partial derivatives of the equations of motion. 
State of the art approaches to this problem often use analytical methods based on the chain rule applied to existing dynamics algorithms. Although these methods are an improvement over finite differences in terms of accuracy, they are not always the most efficient. In this paper, we contribute new closed-form expressions for the first-order partial derivatives of inverse dynamics, leading to a  recursive algorithm. The algorithm is benchmarked against chain-rule approaches in Fortran and against an existing algorithm from the Pinocchio library in C++.  Tests consider computing the partial derivatives of inverse and forward dynamics for robots ranging from kinematic chains to humanoids and quadrupeds. Compared to the previous open-source Pinocchio implementation, our new analytical results uncover a key computational restructuring that enables efficiency gains. Speedups of up to 1.4x are reported for calculating the partial derivatives of inverse dynamics for the 50-dof Talos humanoid.
\end{abstract}

\section{Introduction}

 \begin{table*}[h]
   \centering
  \begin{threeparttable}[b]
   \resizebox{\textwidth}{!}{
    \begin{tabular}{| c | c |c |c |}
    \hline
    \thead{\\ \\ Method} & \thead{Implementation\\ (Easy/Medium/Hard)} & \thead{Speed \\(Slow/Medium/Fast)} & \thead{Accuracy \\ (Low/Medium/High)} \\ \hline
    Finite Difference \cite{tassa,koena,mujoco} & Easy & Medium & Low \\ \hline
     Complex-step Differentiation~\cite{complex,mcx}  & Easy & Slow/Medium & High \\ \hline   
    Automatic Differentiation~\cite{Kudruss,robcogen,drake} & Easy/Medium & Slow/Medium & High \\ \hline
    Analytical Chain-Rule Differentiation~\cite{sohl2001recursive},\cite{car}\tnote{1} & Medium/Hard & Medium & High \\ \hline
    \it{Special-Purpose Analytical Method}~\normalfont \cite{garofalo,jain},\cite{car_code}\tnote{2}, [This Paper]\tnote{3}   & Hard & Medium/Fast  & High \\ \hline
    \end{tabular}}
    \caption{Rough summary of methods to calculate partial derivatives of rigid-body dynamics.}
    \vspace{-3px}
    \label{table1}
 \begin{tablenotes}
    \item[1] Algo-2,3 of \cite{car} provides a chain-rule method for ID partial derivatives.
    \item[2] Pinocchio code for ID partial derivatives accompanying~\cite{car} but different from chain-rule.
    \item[3] This paper provides a {\it Special-Purpose Analytical Method} that outperforms the state-of-the-art in terms of speed.
  \end{tablenotes}    
 \end{threeparttable}
\end{table*}

Rigid-body dynamics models are widely used in the development of control algorithms for quadruped and humanoid robots, with recursive dynamics algorithms at the core of many real-time controllers. Example use cases include current approaches to controller design via optimization~\cite{neunert,posa}. Likewise, robotics libraries such as Crocoddyl~\cite{crocoddyl} and Drake~\cite{drake} require  the partial derivatives of rigid-body dynamics with respect to the state and control variables. The calculation of these derivatives represents the majority of the CPU time required in many optimal control applications~\cite{car}. 

There are multiple general-purpose methods that can be applied to rigid-body dynamics for computing derivatives. Finite differences \cite{tassa,koena,mujoco} are simple to implement and are trivially parallelized, but suffer in accuracy (Table~\ref{table1}). The complex-step method~\cite{cstep} 
is accurate to machine precision but requires all functions to be computed in the complex plane, leading to additional overhead.  Cossette et al.~\cite{complex} extended this method to matrix Lie groups, providing direct applicability to rigid-body dynamics. An approach that avoids the overhead of the complex step is automatic differentiation (\autodiff), also called algorithmic differentiation \cite{Kudruss,robcogen,drake}. \autodiff~calculates the partials of an algorithm by automating the accumulation of chain-rule expressions through operator overloading~\cite{Kudruss} or source code transformation~\cite{drake,robcogen}. The first approach overloads basic data types to compute partials concurrently with all arithmetic. The second approach builds an expression graph from source code and augments the source directly to calculate the partial derivatives.  

Without relying on AD, but similar to it, others have developed analytical methods that accumulate chain-rule expressions on top of existing algorithms. Examples include many existing descriptions of recursive algorithms for dynamics derivatives  \cite{sohl2001recursive,car,lee2005newton}.
The derivation in Ref.~\cite{car} applies chain rule on the classical two-pass Recursive-Newton-Euler Algorithm (RNEA) for inverse dynamics (ID). The presented method has computational complexity $O(Nd)$ for both the forward and backward pass of the RNEA derivatives, where $N$ is the number of bodies in the system and $d$ is the depth of the kinematic connectivity tree. However, the algorithm in the C++ Pinocchio (2.6.0) code~\cite{car_code} associated with \cite{car} is distinctly different from the algorithm presented in \cite{car}. The Pinocchio code includes a more efficient $O(N)$ forward pass, coupled with an $O(Nd)$ backward pass.

Derivative algorithms that directly carry out chain-rule  on an existing algorithm (using AD or by hand) may not always be the most efficient. Other special-purpose methods have been proposed to fully exploit the structure of the rigid-body equations of motion. Such approaches often analytically differentiate closed-form equations of motion and then design an algorithm to compute the result. An example is Ref.~\cite{garofalo} where partial derivatives are considered of the mass matrix $\M(\q)$,  the Coriolis matrix $\C(\q,\qd)$, and the gravity vector $\g(\q)$ with respect to the state variables $\q$ and $\qd$, leading to an $\mathcal{O}(N^3)$ algorithm.

Ref.~\cite{jain} gives a recursive method for partial derivatives of inverse and forward dynamics (FD) for serial kinematic chains with single-DoF joints. Their approach includes an $\mathcal{O}(N^2)$  algorithm for the partials of FD, and an $\mathcal{O}(Nd)$ algorithm for the partials of ID.

The main contribution of this paper is to extend previous analytical results on partial derivatives of ID \cite{jain} to generic multi-DoF joint robots with a fixed or floating base. We first derive closed-form expressions for the first-order partial derivatives of ID. To the best of authors' knowledge, these results represent the first of their kind for general rigid-body systems (fixed or floating base) with multi-DoF joints. The new expressions lead immediately to an algorithm of order $\mathcal{O}(Nd)$ that naturally generalizes the one shown in \cite{jain}. The method is fundamentally different than the straight chain-rule approach presented in \cite{car}. However, the distinct algorithm in the Pinocchio code~\cite{car_code} ends up being directly related to the computations required in our algorithm. Despite the similarity, our new closed-form expressions uncover a key restructuring that accelerates computations.  We use the relationship between FD and ID \cite{car,jain} in an efficient manner to ultimately calculate the partial derivatives of FD with complexity $\mathcal{O}(N^2)$. 
\section{ Derivatives of Rigid-Body Dynamics}
\label{sec:pd_rbd}


\noindent {\bf Rigid-Body Dynamics}: 
 For a rigid-body system, the state variables are the configuration $\q$ and the generalized velocity vector $\qd$, while the control variable is the generalized torque vector $\taubar$. The equation of motion, also called the Inverse Dynamics (ID), is given by 
 \begin{align}
    \taubar &= \M(\q)\qdd+ \C(\q,\qd)\qd + \g(\q)
    \label{inv_dyn} \\
&= ID(model,\q,  \qd, \qdd)
    \label{inv_dyn_model}
\end{align}
where $\M \in \R^{n \times n}$ is the mass matrix, $\C \in \R^{n \times n}$ is a Coriolis matrix, and $\g \in \R^{n}$ is the vector of generalized gravitational forces, and $n$ is the DoF of the system. For a fixed state, ID calculates $\taubar$ for a given $\qdd$ (Eq.~\ref{inv_dyn}), and Forward Dynamics (FD) computes $\qdd$ for a given $\taubar$:
\begin{align}
    \qdd &= \M^{-1}(\q)\left( \taubar -  \C(\q,\qd)\qd - \g(\q) \right)
    \label{fwd_dyn2} \\
    &= FD(model,\q, \qd, \taubar)
    \label{for_dyn_model}
\end{align}

Toward supporting control applications~(e.g., \red{\cite{neunert,koena}}), the objective of this work is to compute the partial derivatives of FD with respect to the state ($\q$,$\qd$) and control variables ($\taubar$). The most efficient algorithms for calculating ID and FD are the  $\mathcal{O}(N)$ Recursive-Newton-Euler Algorithm (RNEA) and the Articulated-Body Algorithm (ABA) respectively~\cite{rbd}.

\vspace{1ex}
{\noindent \bf Derivatives of Inverse and Forward Dynamics:}
As presented in \cite{car}, a direct approach to compute the partial derivatives of ID is to manually differentiate the RNEA algorithm via chain-rule, denoted as RNEACR (Table~\ref{table2}).
A similar chain rule approach for the derivatives of ABA is denoted as ABACR. Because the ABA is more computationally intensive than RNEA, a more efficient approach for derivatives of FD first applies RNEACR, and then uses the relationship \cite{car,jain}
  \begin{equation}
      \frac{\partial~FD}{\partial \boldsymbol u}\biggr\rvert_{\q_{0},  \qd_{0}, \taubar_{0}} = -\M^{-1}(\q_{0}) \frac{\partial~  ID }{\partial \boldsymbol u}\biggr\rvert_{\q_{0},  \qd_{0},  \qdd_{0}}
      \label{car_FO_eqn}
  \end{equation}
  where the variable $ {\boldsymbol u} \in \{ \q$, $\qd \}$. From Eq.~\ref{fwd_dyn2}, $\partial FD / \partial \taubar$
  can be directly calculated as \red{ $\M^{-1}$}, enabling re-use in Eq.~\ref{car_FO_eqn}. This method for the partials of \underline{FD} via RNEA {\underline C}hain {\underline R}ule is abbreviated as FDCR. 

 The state-of-the-art numerical method for computing partial derivatives of FD is provided in the Pinocchio package \cite{car_code} (Pinocchio FD Derivs, Table~\ref{table2}). The method first computes the derivatives of ID (Pinocchio ID Derivs, Table~\ref{table2}). It then computes $\M^{-1}$ and uses Eq.~\ref{car_FO_eqn} to obtain the derivatives of FD. 
 
 In the following sections, Spatial Vector Algebra (SVA) is reviewed for dynamics analysis, followed by a theoretical development of analytical expressions that lead to an algorithm for partial derivatives of ID. The new expressions result in algorithm with a key difference from the state-of-the-art~\cite{car_code}, enabling a speedup over it.


\section{Spatial Vector Algebra (SVA)}
\label{iden_defn}

\makeatletter
\newcommand{\thickhline}{%
    \noalign {\ifnum 0=`}\fi \hrule height 1pt
    \futurelet \reserved@a \@xhline
}
\makeatother

 \begin{table*}[h]
   \centering
   \def\arraystretch{1.5}
    \begin{tabular}{p{2cm}  p{3cm}  p{9cm} }
    \thickhline
   \thead{Quantity} & \thead{Abbreviation} & \thead{Algorithm} \\ \hline \thickhline
   \multirow{3}{*}{$\frac{\partial ID}{\partial \q}$, $\frac{\partial ID}{\partial \qd}$ } & \centering RNEACR  & Forward Accumulation of Chain Rule on RNEA \cite[Algo.~2 \&3]{car}\\ \cline{2-3}
   & \centering Pinocchio ID Derivs & Pinocchio Original Algorithm \cite{car_code}  \\ \cline{2-3}
   & \centering {\bf IDSVA} & Proposed Algorithm using SVA (Algorithm 1) \\ \thickhline
   \multirow{4}{*}{$\frac{\partial FD}{\partial \q}$, $\frac{\partial FD}{\partial \qd}$} & \centering ABACR & Forward Chain Rule on ABA \\ \cline{2-3}
   & \centering FDCR & RNEACR, Compute $\M^{-1}$ \cite{InverseMassMatrix}, Apply Eq.~\ref{car_FO_eqn} \\ \cline{2-3}
   & \centering Pinocchio FD Derivs & Pinocchio ID Derivs, Compute $\M^{-1}$ \cite{InverseMassMatrix}, Apply Eq.~\ref{car_FO_eqn} \\ \cline{2-3}
   & \centering {\bf FDSVA} & IDSVA, Compute $\M^{-1}$ \cite{InverseMassMatrix}, use DMM/AZA depending on $N$ for Eq.~\ref{car_FO_eqn}\\ \thickhline
   $ABA(\q,0,\b,0)$ & \centering \bf{AZA} & Simplified ABA with select zero inputs for Eq.~\ref{car_FO_eqn}\\ \thickhline
    \end{tabular}
    \caption{Abbreviations of various algorithms/methods used. The bold acronyms are contributions from the current paper.}
    \label{table2}
\end{table*}

\noindent {\bf Notation:} Spatial vectors are 6D vectors that combine the linear and angular aspects of a rigid-body motion or net force~\cite{rbd}. Spatial vectors are denoted with  lower-case bold letters (e.g., $\a$), while matrices are denoted with capitalized bold letters (e.g., $\boldsymbol{A}$). Motion vectors, such as velocity and acceleration, belong to a 6D vector space denoted $M^{6}$. Spatial vectors are usually expressed in either the ground coordinate frame or a body coordinate (local coordinate) frame. For example, the spatial velocity ${}^{k}\v_{k} \in M^{6}$ of a body $k$ expressed in the body frame is given by
${}^{k}\v_{k} = \begin{bmatrix}
           {}^{k}\omega_{k}^T &
           {}^{k}v_{k}^T
         \end{bmatrix}^T$
where ${}^{k}\omega_{k} \in \mathbb{R}^{3}$ is the angular velocity expressed in a coordinate frame fixed to the body, while ${}^{k}v_{k} \in \mathbb{R}^{3}$ is the linear velocity of the origin of the body frame. When expressed in the ground frame, the spatial velocity for body $k$ is denoted as ${}^{0}\v_{k}$. The vector is again composed of angular and linear velocity components. However, the linear velocity is associated with the body-fixed point on body $k$ that is coincident with the origin of the ground frame $0$. In this paper, when the frame used to express a spatial vector is omitted, the ground frame is assumed.


Force-like vectors, such as force and momentum, belong to another 6D vector space $F^{6}$, which is dual to $M^{6}$. A spatial cross-product between two motion vectors ($\v$,$\u$), written as $(\v \times) \u$, is given by (Eq.~\ref{cross_defn_1}). This operation can be understood as providing the time rate of change of $\u$, when $\u$ is moving with a spatial velocity $\v$. 
For a Cartesian vector, the matrix $\omega \times$ is the classical 3D cross product operator. A spatial cross-product between a motion and a force vector is written as $(\v \times^*) \f$, as defined in Eq.~\ref{cross_f_defn_1}.
\begin{multicols}{2}
  \begin{align}
    \v \times = \begin{bmatrix}
        \omega \times & \bf{0}  \\
       v \times &  \omega \times
    \end{bmatrix}
    \label{cross_defn_1}
\end{align}
  \begin{align}
    \v \times^* = \begin{bmatrix}
        \omega \times & v \times  \\
        \bf{0} &  \omega \times
    \end{bmatrix}
    \label{cross_f_defn_1}
\end{align}
\end{multicols}
\vspace{-1.5ex}
\noindent An operator $\crff$ is  defined by swapping the order of the cross product, such that $(\f \crff) \v = (\v \times^*) \f$~\cite{eche}. Further introduction to  SVA is provided in \blue{Appendix~\ref{sva_intro}}.

\begin{figure}[tb]
\centering
\includegraphics[width=.85 \columnwidth]{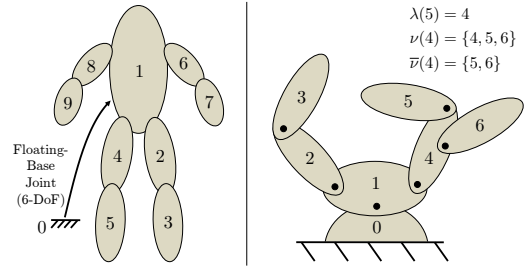}
\caption{Body numbering and notation examples with floating- and fixed-base systems.}
\label{link_fig}
\end{figure}

\vspace{1ex}
\noindent {\bf Connectivity:} An open-chain kinematic tree with serial or branched connectivity (Fig.~\ref{link_fig}) is considered with $N$ links connected by joints, each with up to 6 DoF. Body $i$'s parent toward the root of the kinematic tree is denoted as $\lambda(i)$. $\nu(i)$ denotes the set of bodies in the subtree rooted at body $i$, while $\subtreeb(i)$ denotes the set of bodies in $\subtree(i)$ excluding the body $i$ . We define $i \preceq j$ if body $i$ is in the path from body $j$ to the base. 

The spatial velocities of the neighbouring bodies in the tree are related by $\v_{i} = \v_{\lambda(i)} + \Phibar_{i} \qd_{i}$, where $\Phibar_{i}$ is the joint motion subspace matrix for joint $i$ \cite{rbd} and $\qd_{i}$ the joint rates for joint $i$. For a single DoF revolute joint, $\Phibar_{i} \in \mathbb{R}^{6}$, and $\qd_{i}$ is a scalar. For a 6-DoF free-motion joint, $\Phibar_{i} \in \mathbb{R}^{6 \times 6}$,  while $\qd_{i} \in \mathbb{R}^{6}$. The velocity $\v_{i}$ can also be written as the sum of joint velocities over predecessors as $\v_{i} = \sum_{l \preceq i} \red{\Phibar_{l}} \qd_{l}$. 
 The derivative of joint motion \red{subspace matrix} due to the axis changing with respect to local coordinates (commonly denoted \red {$\mathring{\Phibar}_{i}$}~\cite{rbd}) is assumed to be zero.
The quantity $\red{\Phibardot{}_{i}} = \v_{i} \times \red{\Phibar_{i}}$ signifies the rate of change of $\Phibar_{i}$ due to the local coordinate system moving.

\vspace{1ex}
\noindent {\bf Dynamics:} The spatial equation of motion~\cite{rbd} is given for body $k$ as:
\begin{equation}
    \f_{k}= \I_{k}\a_{k} + \v_{k} \times^*\I_{k} \v_{k}
    \label{eq:spatial_eom}
\end{equation}
where $\f_{k}$ is the net spatial force on body $k$, $\I_{k}$ is its spatial inertia~\cite{rbd}, and $\a_{k}$ is its spatial acceleration. Instead of treating gravity as an external force, a common trick is to accelerate the base upwards opposite of the gravitational acceleration ($\a_{0} = -\a_{g}$), providing the acceleration of body $k$ as:
\begin{equation}
    \a_{k} = \textstyle\sum_{l \preceq k} \red{\big(} \red{\Phibar_{l}} \qdd_{l} + \v_{l} \times \red{\Phibar_{l}} \qd_{l} \red{\big)}+\a_{0}\,.
    \label{spatial_acc}
\end{equation}
For later use, we decompose \red{$\a_k$} into terms from joint accelerations, and terms from joint rates, according to
\[
 \gammabar_{k} = \textstyle \sum_{l \preceq k} \red{\Phibar_{l}} \qdd_{l}
\textrm{~~~and~~~}
\xibar_{k} = \textstyle \sum_{l \preceq k} \v_{l} \times \red{\Phibar_{l}} \qd_{l}  
\]
\red{($\gammabar, \xibar \in M^{6}$)}. With these definitions, \red{$\a_k = \gammabar_k+\xibar_k + \a_{0}$}.

In a similar fashion, the net spatial force on body $k$ is then decomposed as:
\begin{equation}
    \f_{k}=\etabar_{k} + \zetabar_{k}+\I_{k}\a_{0}
    \label{f_defn}
\end{equation}
where $\etabar_{k} = \I_{k} \gammabar_{k}$ \red{($\etabar \in F^{6}$)} is the spatial force on the body caused by joint accelerations $\qdd$, and  $\zetabar_{k} = \v_{k} \times^{*}\I_{k} \v_{k} + \I_{k} \xibar_{k}$ ($\zetabar \red{\in F^{6}}$) gives the Coriolis and centripetal forces on body $k$. From the formulation of the RNEA~\cite{rbd}, $\red{\taubar_{i}} = \Phibar_{i}^{T}\f_{i}^{C} $, where $\f_{i}^{C}=\sum_{k \succeq i} \f_{k}$ is the spatial force transmitted across joint $i$.

\begin{figure*}[!b]
\hrulefill
\begin{align}
     \frac{\partial [\C\qd]_{i}}{\partial \red{\q_{j}}} = & \red{\Phibar_{i}^{T}  \Big[  2 \B_{i}^{C} \Psibardot_{j}  +     \I_{i}^{C} \v_{\lambda(j)}  \times \Psibardot_{j}   + } 
 \red{\I_{i}^{C} \xibar_{\lambda(j)} \times \Phibar_{j}  \Big] } 
     \label{CFO_q_eqn_1}
 \tag{26}\\
     \frac{\partial [\C\qd]_{j}}{\partial \red{\q_{i}}} = & \red{\Phibar_{j}^{T}  \Big[  2 \B_{i}^{C} \Psibardot_{i}  +     \I_{i}^{C} \v_{\lambda(i)} \times \Psibardot_{i} +}
     \red{\I_{i}^{C} \xibar_{\lambda(i)} \times \Phibar_{i}  + \zetabar_{i}^{C} \crff \Phibar_{i} \Big] , (j \neq i) }
    \label{CFO_q_eqn_2}
    \tag{27}
\end{align}
\end{figure*}

Proceeding to consider the system overall, for any particular joint $i$, Eq.\red{~\ref{inv_dyn}} can be written as follows.

\begin{equation}
    \red{\taubar_{i}} =   [ \M(\q) \qdd]_{i}+   [\C(\q,\qd) \qd]_{i}  + \red{\g_{i}}(\q)
    \label{inv_dyn_expanded_eq2}
\end{equation}
Then, using $\red{\taubar_{i}}=\Phibar_{i}^{T}\f_{i}^{C} $, Eq.~\red{\ref{spatial_acc}} and \red{\ref{inv_dyn_expanded_eq2}}: 
\begin{align}
 [\, \M(\q) \qdd \,]_{i} &= \red{\Phibar_{i}^{T}} \sum_{k \succeq i} [\I_{k} \gammabar_{k}]  
\label{mqdotdot_exp_2}\\[.5ex]
   [\,\C(\q,\qd) \qd\,]_{i} &= \red{\Phibar_{i}^{T}} \sum_{k \succeq i}[ \v_{k} \times^* \I_{k} \v_{k} + \I_{k}  \xibar_{k} ]
    \label{cqdot_exp_2}\\[.5ex]
    \red{\g_{i}} &=
   \red{ \Phibar_{i}^{T} \I_{i}^C  \a_{0}}
   \label{g_eqn}
\end{align}
where $\I{}_{i}^C$ denotes the composite rigid-body inertia of the sub-tree rooted at body $i$, given by $\I{}_{i}^{C} = \sum_{k \succeq i} \I_{k}$.

\section{Analytical Partial Derivatives using SVA}
\label{sec:fdsva}

In this section, closed-form expressions are derived for the derivatives of ID. 
The reader less interested in the derivation may skip to Eq.~\ref{IDFO_SVA_eq1} and Eq.~\ref{Cqdot_FO_qdot_eqn} as a summary.

\newcommand{\qjp}[2]{q_{#1,#2}}

\subsection{Building Blocks}
Several kinematic identities are given in \blue{Appendix~\ref{multi_dof_iden}}  as a basis for the main derivation.
 \red{We denote by $n_j$ the number of DoFs for joint $j$. The  motion subspace matrix for the joint is then given as
 $
 \Phibar_j = \begin{bmatrix} \phibar_{j,1} & \cdots \phibar_{j,n_j} \end{bmatrix}
 $
 where each spatial vector $\phibar_{j,p}$ gives the $p$-th free-mode of motion for joint $j$.
 With a slight liberty of notation, $\partial/ \partial \qjp{j}{p}$ denotes an operator for the directional derivative along this $p$-th free mode of a joint. For example, considering the case with $j \preceq i$ gives identity J1: 
 \begin{equation}
 \frac{\partial }{\partial \qjp{j}{p}} \Phibar_i = \phibar_{j,p}\times \Phibar_i \tag{J1}
 \end{equation}
 which provides the rate of change in $\Phibar_i$ with respect to relative motion $\phibar_{j,p}$ at joint $j$ earlier in the chain. For revolute joints, these derivatives are just conventional derivatives with respect to joint angles. For multi-DoF joints such as a floating base, they are more formally Lie derivatives. 
Considering a configuration-dependent vector $\boldsymbol{u}$, we denote by
\[
\frac{\partial \boldsymbol u}{\partial \q_j} = \Large \begin{bmatrix}\nicefrac{\partial \boldsymbol u}{\partial \qjp{j}{1}} & \cdots & \nicefrac{\partial \boldsymbol u}{\partial \qjp{j}{n_j}} \end{bmatrix}
\]
the matrix of derivatives associated with joint $j$. To illustrate, we derive identity J7 for later use in the section.


Considering $j \preceq i$ and using the definition of $\gammabar_{i}$:} 
\begin{equation}
    \frac{\partial \gammabar_{i}}{\partial \qjp{j}{p}} = \sum_{l \preceq i} \frac{\partial \Phibar_{l}}{\partial \qjp{j}{p}}  \qdd_{l}
    \label{dgamma_eq1}
\end{equation}
Using J1 and switching the order of the cross product:

\begin{equation}
    \frac{\partial \gammabar_{i}}{\partial \qjp{j}{p}} = - \sum_{j \preceq l \preceq i} \Phibar_{l} \qdd_{l}  \times \phibar_{j,p}
\label{dgamma_eq3}
\end{equation}

\noindent Collecting all DoFs of joint $j$, Eq.~\ref{dgamma_eq3} becomes:

\begin{equation}
    \frac{\partial \gammabar_{i}}{\partial \q_{j}} =  - \sum_{j \preceq l \preceq i} \Phibar_{l} \qdd_{l}  \times \Phibar_{j}
\label{dgamma_eq4}
\end{equation}

\noindent Using the definition of $\gammabar_{l}$, and summing over $l$:

\begin{equation}
    \frac{\partial \gammabar_{i}}{\partial \q_{j}} = (\gammabar_{\lambda(j)} - \gammabar_{i}  )  \times \Phibar_{j}
\label{dgamma_eq5}
\end{equation}

\subsection{First-order Partial Derivatives of ID w.r.t.~$\q$}

\label{FO_ID_adsva_sec}

The subsequent derivations employ these formulae and the identities from \blue{Appendix \ref{multi_dof_iden}} to obtain the partials of the terms in Eq.~\red{\ref{inv_dyn_expanded_eq2}}. Since the derivations are quite lengthy, the presentation focuses on explaining the methodology. A full derivation is provided in \blue{Appendix~\ref{partials_details}}. 

\vspace{1ex}
{\noindent \bf Partial Derivative of Eq.\red{~\ref{mqdotdot_exp_2}}:} We derive the partial derivative of $[ \M(\q) \qdd]_{i}$ with respect to \red{$\q_{j}$} for all $ i,j \in \{1,\ldots,N\}$. First, consider the case when $j \preceq i$. Using the product rule of differentiation in Eq.~\ref{mqdotdot_exp_2}:
  \begin{equation}
      \begin{aligned}
        \frac{\partial [ \M(\q)\qdd]_{i}}{\partial \q_{j}} = & \frac{\partial (\Phibar_{i}^{T})}{\partial \q_{j}} \sum_{k \succeq i} [\I_{k} \gammabar_{k}]   +  \\ &~~~    \Phibar_{i}^{T}  \sum_{k \succeq i} \Bigg[\frac{\partial \I_{k}}{\partial \q_{j}} \gammabar_{k} +  \I_{k} \frac{\partial \gammabar_{k}}{\partial \q_{j}} \Bigg]  
      \end{aligned}
\end{equation}
  Using identity J3 for the term $\frac{\partial (\Phibar_{i}^{T})}{\partial \q_{j}}$, J4 for the term $\frac{\partial \I_{k}}{\partial \q_{j}} \gammabar_{k}$, J7 for $\frac{\partial \gammabar_{k}}{\partial \q_{j}}$, and cancelling terms:
  

  \begin{equation}
      \begin{aligned}
         \frac{\partial [ \M(\q) \qdd]_{i}}{\partial \q_{j}} =  &  \Phibar_{i}^{T}  \sum_{k \succeq i} \Big[
         \I_{k}  ( \gammabar_{\lambda(j)} \times ) \Big]   \Phibar_{j}
      \end{aligned}
\end{equation} 

\noindent Upon summing over the index $k$, the final expression is:

\begin{equation}
      \frac{\partial [ \M(\q) \qdd]_{i}}{\partial \red{\q_{j}}} =  \red{\Phibar_{i}^{T} \Big[  \I_{i}^{C}  \gammabar_{\lambda(j)}   \times \Phibar_{j} \Big]} 
    \label{partial_Mqddot_1}
\end{equation}
For the case $i \red{\prec} j$, we have that:
 \begin{equation}
         \frac{\partial [ \M(\q) \qdd]_{i}}{\partial \q_{j}} =   \Phibar_{i}^{T}  \sum_{k \succeq i} \Bigg[\frac{\partial \I_{k}}{\partial \q_{j}} \gammabar_{k} + \I_{k} \frac{\partial \gammabar_{k}}{\partial \q_{j}} \Bigg]
        \label{partial_Mqddot_4}
\end{equation}
 
\noindent Using the identities \red{J3, J4 and J7} and cancelling terms:
   \begin{equation}
      \begin{aligned}
       \frac{\partial [ \M(\q) \qdd]_{i}}{\partial \q_{j}} = 
     \Phibar_{i}^{T}  \sum_{ k \succeq j} \Big[(\I_{k} \gammabar_{k}) \crff + \I_{k}  (\gammabar_{\lambda(j)}  \times)  \Big]\Phibar_{j}  
        \label{partial_Mqddot_5}
      \end{aligned}
\end{equation} 

Summing over the index $k$ results in:

\begin{equation}
\frac{\partial [ \M(\q) \qdd]_{i}}{\partial \red{\q_{j}}} = \red{\Phibar_{i}^{T} \Big[\etabar_{j}^{C} \crff  + \I_{j}^{C} (\gammabar_{\lambda(j)} \times) \Big] \Phibar_{j}}
    \label{partial_Mqddot_2}
\end{equation}
where  $\etabar_{j}^{C}$ collects joint-acceleration-dependent forces for the subtree, calculated as $\etabar_{j}^{C} = \sum_{k \succeq j} \etabar_{k}$ .

For ease of implementation, we consider a single case where $j \preceq i$. The indices $i$ and $j$ are switched in Eq.\red{~\ref{partial_Mqddot_2}}, to get the expression for $\frac{\partial [\M(\q)\qdd]_{j}}{\partial \q_{i}}$ for the case \red{$j \prec i$}. Therefore,  Eq.\red{~\ref{Mqddot_FO_eqn}} gives the two expressions formulated for the general case $j \preceq i$.

\begin{align}
    & \frac{\partial [ \M(\q) \qdd]_{i}}{\partial \red{\q_{j}}} = \red{\Phibar_{i}^{T} \Big[  \I_{i}^{C}  \gammabar_{\lambda(j)}   \times \Phibar_{j} \Big]}  \label{Mqddot_FO_eqn} \\
    & \frac{\partial [ \M(\q)\qdd]_{j}}{\partial \red{\q_{i}}} = \red{\Phibar_{j}^{T} \Big[\etabar_{i}^{C} \crff + \I_{i}^{C} \gammabar_{\lambda(i)} \times  \Big] \Phibar_{i}, (j \neq i)} \nonumber
\end{align}

\vspace{1ex}
{\noindent \bf 
Partial Derivative of Eq.\red{~\ref{cqdot_exp_2}}:} Similarly, we use identities \red{J2-J6 (\blue{Appendix \ref{multi_dof_iden})}} to get the first-order partial derivatives of $[\C(\q,\qd) \qd]_{i}$ with respect to \red{$\q_{j}$} for the case  $j \preceq i$ as shown in Eq.~\ref{CFO_q_eqn_1},~\ref{CFO_q_eqn_2}.
\stepcounter{equation}
\stepcounter{equation}

 In these equations, the composite of $\zetabar_{i}^C$ for the subtree is defined as $\zetabar{}_{i}^{C} = \sum_{k \succeq i} \zetabar_{k}$, and the matrix $\B_{k}(\v_{k},\I_{k})$ is a body-level Coriolis matrix \cite{nei_ms,nei_slotine,eche} given by:
\begin{equation}
    \B_{k} = \frac{1}{2} \big[ \big(\v_{k}\times^* \big)\I_{k} - \I_{k} \big(\v_{k} \times   \big) + \big(\I_{k} \v_{k} \big) \crff    \big]
    \label{bl_term_defn}
\end{equation}
with its composite $\B{}_i^C = \sum_{k\succeq i} \B_k$. 
\vspace{1ex}
{\noindent \bf Partial Derivative of the Gravity Term:} \red{
Using the identities J3 and J4,} for the case $j \preceq i$, the partial derivative of \red{$\g_{i}$} (Eq.~\ref{g_eqn}) with respect to \red{$\q_{j}$} is:
\begin{equation}
    \begin{aligned}
         &  \frac{\partial \red{\g_{i}}}{ \partial \red{\q_{j}}} = \red{\Phibar_{i}^{T}  \I_{i}^{C}(\a_{0} \times \Phibar_{j})} \\
        & \frac{\partial \red{\g_{j}}}{ \partial \red{\q_{i}}} = \red{\Phibar_{j}^{T} \Big[(\I_{i}^{C}\a_{0})\crff + \I_{i}^{C}(\a_{0} \times ) \Big]\Phibar_{i} , (j \neq i)}
        \label{gFO_q_eqn}
    \end{aligned}
\end{equation}

\noindent {\bf Summary for Partials of ID w.r.t.~$\q$:} The partials of the individual components are now collected together. 

\noindent Terms in Eqs.\red{~\ref{Mqddot_FO_eqn}}, \red{\ref{CFO_q_eqn_1}, \ref{CFO_q_eqn_2}},  and \red{\ref{gFO_q_eqn}} are added to get the total expressions for $\frac{\partial \taubar}{\partial \q}$ for the case $j \preceq i$. \red{The full derivation is provided in \blue{Appendix~\ref{combine_terms}}}.

\begin{align}
        \frac{\partial \red{\taubar_{i}}}{\partial \red{\q_{j}}} = & \red{ \Phibar_{i}^{T} \big[ 2  \B_{i}^{C} \big] \Psibardot{}_{j} + \Phibar_{i}^{T}  \I_{i}^{C} \Psibarddot{}_{j} } \label{IDFO_SVA_eq1} \\
        \frac{\partial \red{\taubar_{j}}}{\partial \red{\q_{i}}} = & \red{\Phibar_{j}^{T} [ 2 \B_{i}^{C} \Psibardot_{i} +  \I_{i}^{C}  \Psibarddot_{i}+(\f_{i}^{C} )\crff \Phibar_{i}  ] (j \neq i)} \nonumber
\end{align}
\red{The spatial quantities $\Psibardot_{j}$ and $\Psibarddot_{j}$ are defined as:}

\begin{align}
\begin{split}
  \red{\Psibardot_{j} = }& \red{\v_{\lambda(j)} \times \Phibar_j} \\
\red{ \Psibarddot_{j} = }& \red{\a_{\lambda(j)} \times \Phibar_{j} + \v_{\lambda(j)} \times \Psibardot_{j}}
\end{split}
\label{psidot_psidotdot_defn}
\end{align}

\subsection{First-Order Partial Derivatives of ID w.r.t.~$\qd$}

The partials of $\taubar$ with respect to $\qd$ \red{depend only on the the Coriolis terms $\C \qd$}. Using the identities \red{J8 and J9} leads to expressions for the case when $j \preceq i$(see  \blue{Appendix~\ref{partials_qd}} for full derivation):

\begin{equation}
    \begin{aligned}
    & \frac{\partial \red{\taubar_{i}}}{\partial \red{\dot{\q}_{j}}} =\frac{\partial [\C  \qd]_{i}}{ \partial \red{\dot{\q}_{j}}} = \red{\Phibar_{i}^{T} \Big[2 \B_{i}^{C} \Phibar_{j} +  \I_{i}^{C} (    \Psibardot_{j} + \Phibardot_{j} ) \Big]  }\\
    & \frac{\partial \red{\taubar_{j}}}{\partial \red{\dot{\q}_{i}}} =\frac{\partial [\C  \qd]_{j}}{ \partial \red{\dot{\q}_{i}}} = \red{\Phibar_{j}^{T} \Big[2 \B_{i}^{C} \Phibar_{i} +  \I_{i}^{C} (    \Psibardot_{i} + \Phibardot_{i} ) \Big]  (j \neq i)}
\label{Cqdot_FO_qdot_eqn}
    \end{aligned}
\end{equation}

\subsection{Algorithm for First-Order Partials of ID}
Algorithm \ref{alg:tau_FO_v2} returns the terms in Eq.\red{~\ref{IDFO_SVA_eq1}} and Eq.\red{~\ref{Cqdot_FO_qdot_eqn}} and has computational complexity $\mathcal{O}(Nd)$. The method operates with all spatial vectors expressed in ground frame coordinates. 
The first pass is in the forward direction and goes from root to leaves calculating the spatial velocity, acceleration, $\Phibardot_{i}$, $\Psibardot_{i}$, $\Psibarddot_{i}$, $\B_{i}$ and the spatial force $\f_{i}$. 
When joint $i$ has a single DoF, $\Phibardot_i = \Psibardot_i$ and $\ddot{\Phibar}_i =\Psibarddot_{i}$ and the steps are the same as in \cite[Alg.~2]{jain}.

The second pass progresses in the backward direction from leaves to the root. Index $i$ takes all the values from 1 to $N$. The quantity $\frac{\partial \boldsymbol\tau}{\partial \q}[\subtree(i), i]$ denotes the partial derivatives of $\boldsymbol\tau$ for all bodies in the sub-tree of $i$ with respect to $\q_{i}$, while $\frac{\partial \boldsymbol\tau}{\partial \q}[i, \subtreeb(i)]$ is the partial derivative of $\boldsymbol\tau$ for body $i$ with respect to each of the bodies in the sub-tree of $i$, excluding body $i$. 
This backward pass again generalizes the one in \cite[Alg.~2]{jain}. The differences are that the third term on line 15 can be dropped in the single-DoF case (since $ {\Phibar}_i^T (\f_i^C \crff) {\Phibar}_i =0$ in that case), while lines 17-20 restructure a second inner loop in \cite{jain} into matrix multiplies. 

\begin{algorithm}[t]
\small
\caption{IDSVA Algorithm}

\begin{algorithmic}[1]  
\REQUIRE $ \q, \,\qd ,\, \qdd,\, model$

\STATE $ \v_0 = 0; \, \a_{{0}} = -\a_g  $

\FOR{$i=1$ to $N$}


\STATE  $ \v_{i} =  \v_{\lambda(i)} +  \red{\Phibar_i} \qd_{i}$


\STATE $\a_{i} =  \a_{\lambda(i)} +  \Phibar_i \ddot{\q}_{i} + \v_i \times \Phibar_{i} \qd_i$ \\

\STATE $ \red{\Phibardot_i}  =  \v_{i} \times \red{\Phibar_i}  $\\[.5ex] 

\red{\STATE $\Psibardot_i  =  \v_{\lambda(i)}\times \Phibar_i  $\\[.5ex] 

\STATE $\Psibarddot_i  =  \a_{\lambda(i)}\times \Phibar_i + \v_{\lambda(i)}\times \Psibardot_i   $ \\[.5ex]

}

\STATE $\I_i^C = \I_i$
\STATE $ \B_i^C = \frac{1}{2}[ ( \v_{i} \times^*) \I_i -  \I_i  ( \v_{i} \times ) + (\I_i \v_i )\crff ]  $\\[.5ex]

\STATE $ \f_i^C =  \I_i  \a_{i} + ( \v_{i} \times^*) \I_i \v_{i}  $\\[.5ex]  

\ENDFOR


\FOR{$i=N$ to $1$}

\STATE $ \t_1[i] = \I{}_{i}^C  \red{{\Phibar}_i} $\\[.5ex] 

\STATE $\t_2[i] = 2\B{}_i^C \red{\Phibar_i} + \I_i^C ( \red{\Phibardot_i + \Psibardot_i } )$\\[.5ex] 

 \STATE $\t_3[i] = 2 \B{}_i^C  \red{\Psibardot_i} + \I{}_{i}^C \red{\Psibarddot_i}  + \red{\f_i^C \crff {\Phibar}_i}  $\\[.5ex]
\STATE $\t_4[i] =   2[\B_i^C]\T \red{{\Phibar}_i}$\\[.5ex] 

\STATE $\frac{\partial \boldsymbol\tau}{\partial \q}[i, \subtreeb(i)] = \Phibar_i\T \t_3[\subtreeb(i)]$ \\[1ex]

\STATE $\frac{\partial \boldsymbol\tau}{\partial \qd}[i, \subtreeb(i)] = \Phibar_i\T \t_2[\subtreeb(i)]$ \\[1ex]

\STATE $\frac{\partial \boldsymbol\tau}{\partial \q}[\subtree(i), i] =\t_4[\subtree(i)]\T \Psibardot_i +  \t_1[\subtree(i)]\T \Psibarddot_i $ \\[1ex]

\STATE $\frac{\partial \boldsymbol\tau}{\partial \qd}[ \subtree(i), i] = \t_4[\subtree(i)]\T \Phibar_i + \t_1[\subtree(i)]\T (\Phibardot_i + \Psibardot_i) $ \\[1ex]

    \IF {$\lambda(i) > 0$}
    
        \STATE $\I_{\lambda(i)}^C = \I_{\lambda(i)}^C + \I_i^C ; \, \B_{\lambda(i)}^C = \B_{\lambda(i)}^C +  \B_i^C $ \\[.5ex]
        \STATE $ \f_{\lambda(i)}^C = \f_{\lambda(i)}^C +  \f_i^C $ 

    \ENDIF

\ENDFOR
\RETURN $\frac{\partial \taubar }{\partial{\q}},\frac{\partial \taubar }{\partial{\qd}}$

\end{algorithmic}

\label{alg:tau_FO_v2}
\end{algorithm}


\subsection{Relating Partials of FD and ID}
\label{sec:efficient_SVA}

Equation~\ref{car_FO_eqn}  gives the relation between the derivative of FD and ID, where $\frac{\partial ID }{\partial \u}$ ($\u = [\q, \qd ]$) is an $n \times 2n$ matrix, and $\M^{-1}$ is an $n \times n$ matrix. A direct multiplication of the two matrices results into an $\mathcal{O}(N^3)$ operation and is named Direct Matrix Multiplication (DMM). An alternative method with reduced computational complexity ($\mathcal{O}(N^2)$) 
is presented in this section. 

ABA gives $\qdd$ (Eq.~\ref{fwd_dyn2}) for a given $\taubar$, represented as: 
\begin{equation}
       \qdd = ABA(\q,\qd,\taubar,\a_g)
    \label{ABA_eqn_1}
\end{equation}
From Eq.~\ref{fwd_dyn2}, for $\qd=\bf{0}$, $\a_g=\bf{0}$ and an arbitrary input vector $\taubar=\b$,
$
     \qdd = \M^{-1} \b
    \label{fwd_dyn3}
    $.
The product of the matrix $\M^{-1}$ with any vector $\b$ can therefore be calculated by: 
\begin{equation}
       \M^{-1} \b  = ABA(\q,\bf{0},\b, \bf{0})
    \label{ABA_eqn_2}
\end{equation}
For a product of $\M^{-1}$ with any given matrix (of size $n \times m$), ABA can be used $m$ times with $\qd=\bf{0}$, $\a_g=\bf{0}$ and $\b$ as the column vectors of the given matrix one at a time, resulting in an $\mathcal{O}(Nm)$ operation. Due to the repeated use of ABA for each input column vector, the kinematic variables and articulated inertias are only calculated once, and saved for re-use. To implement Eq.~\ref{car_FO_eqn},  Eq.\red{~\ref{ABA_eqn_2}} is used with $\b$ as the column vectors of $\frac{\partial ID }{\partial \u}$ one at a time. This process is defined the ``ABA-Zero-Algorithm'' (AZA), since inputs $\qd$ and $\a_g$ are $\bf{0}$. 
\section{Results}

\subsection{Algorithm correctness}
 The partial derivatives of FD from ABACR, FDSVA, \& FDCR  are compared with derivatives calculated \red{in a Fortran implementation} using the complex-step method for accuracy. For $N=100$, the ABACR method results in a term-by-term root-mean-square (rms) relative error of $10^{-13}$, while both FDSVA and FDCR result in an rms error of approximately $10^{-12}$. The rms relative error for all methods grows linearly with DoF on a log-log scale. 

\subsection{Runtime for Partials of ID/FD vs. RNEACR/FDCR as presented in \cite{car} via Fortran implementation of both}

We consider an $N$ link serial \red{or branched} kinematic tree with all revolute joints about their local $z$-axis. 

\begin{figure}[tb]
\hspace{-0.5cm}
\includegraphics[width=9cm]{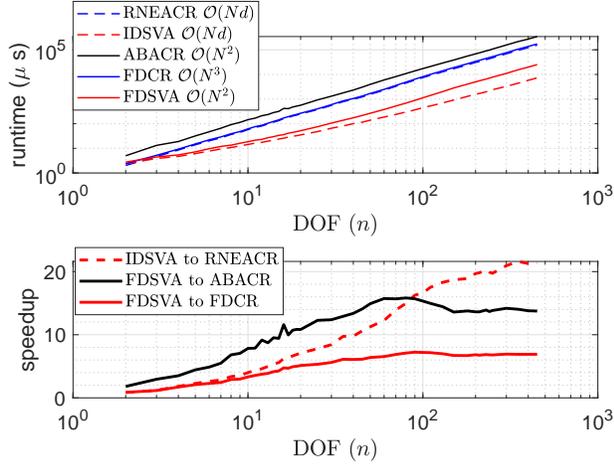}
\caption{Fortran implementation of IDSVA outperforms RNEACR, FDSVA outperforms FDCR and ABACR (Table~\ref{table2}) for serial chains ($N = n$ for revolute joints) with all $N \succeq 2$}
\label{all_plots_fortran}
\end{figure}

RNEACR in body-coordinates\cite[Algos.~2 \& 3]{car} (Table~\ref{table2}) is used to calculate the partial derivatives of ID, and compared with the IDSVA (Table~\ref{table2}) method. \red{All the} algorithms are written in Fortran 90 and implemented using the Intel Fortran compiler on a 3.07 GHz Intel Xeon processor. To calculate the average run time, each algorithm is run 10,000 times with randomized inputs for the  state and control variables. Fig.~\ref{all_plots_fortran} shows the comparison of the two methods. For $N=100$, a speedup of $15 \times$ for IDSVA over RNEACR is found. 

 \red{Fig.~\ref{all_plots_fortran} also shows the comparison of FDSVA with FDCR and ABACR (Table~\ref{table2}) for serial chains.} With the analytical derivative expressions developed herein, FDSVA method outperforms FDCR for all values of $N \ge 2$. 
 
\begin{figure}[tb]
\hspace{-0.5cm}
\includegraphics[width=9cm]{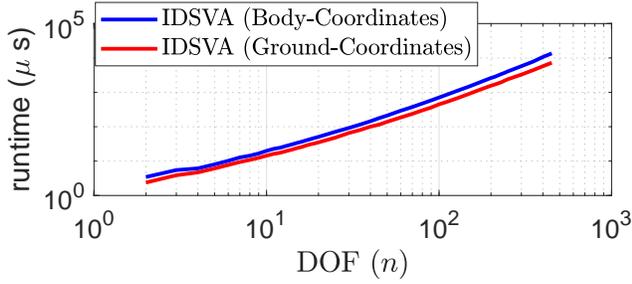}
\caption{\blue{Fortran implementation of IDSVA in ground vs body coordinate frame.}}
\label{idsva_bf_vs_gf}
\end{figure}

 \blue{Since SVA allows for coordinate-free expressions, recursive algorithms can be formulated in either body-coordinates or ground-coordinates~\cite{rbd}. For the body-coordinate algorithms, all the intermediate quantities are transformed between body coordinates using transformation matrices $\prescript{i}{}{\boldsymbol{X}}_{\lambda(i)}$, while for the ground-coordinates algorithms, the quantities can be left in the ground frame. A major advantage for the latter comes by avoiding the repeated transformation of the quantities between local body frames in the backward pass of the algorithm. This gain is achieved at the cost of expressing kinematic quantities (velocities $\prescript{0}{}{\v}_{i}$, accelerations  $\prescript{0}{}{\a}_{i}$), joint motion subspace matrices  $\prescript{0}{}{\Phibar}_{i}$, and inertia matrices  $\prescript{0}{}{\I}_{i}$ in the ground coordinate frame during the outward pass.  Fig.~\ref{idsva_bf_vs_gf} shows a comparison of IDSVA (see Table~\ref{table2}) runtime in body coordinates and ground coordinates. Speedups for ground coordinate algorithms are between 1.3 to 1.9 for $N=2$ to $N=500$. 
 }

\begin{figure}[tb]
\includegraphics[width=9cm]{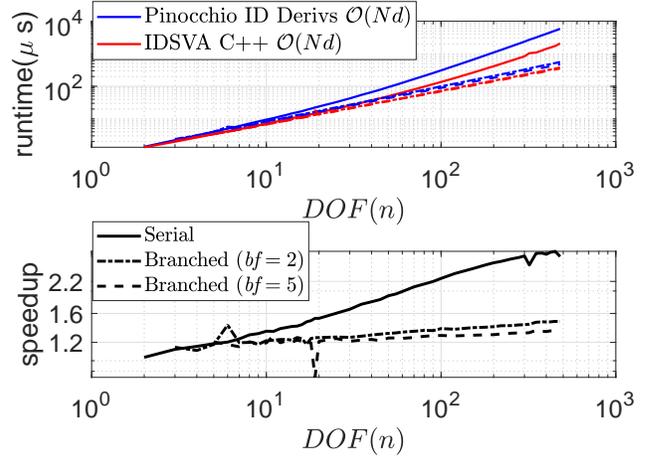}
\caption{a) Serial chain (solid), Branched chain (\textit{bf}=2, dashed dotted), Branched chain (\textit{bf}=5, dashed) b) \red{IDSVA C++ improves upon Pinocchio ID Derivs for all $N \geq 2$ (gcc-9.0 compiler). $N = n$ for kinematic trees with revolute joints.}}
\label{SVA_vs_RNEA_FO_fig}
\end{figure}

\subsection{Comparing Runtime for C++ Partials of ID/FD with Pinocchio Implementation Accompanying \cite{car}}
\begin{figure}[tb]
\hspace{-0.5cm}
\center
\includegraphics[width=8.5cm]{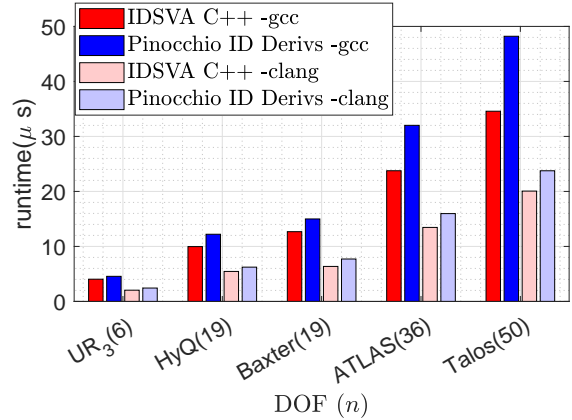}
\caption{\red{Comparison of IDSVA (in C++) with Pinocchio ID Derivs~\cite{car_code} framework for several floating base multi-dof robots ($n$)- Fixed base- UR3, Baxter. Floating base- HyQ, ATLAS, Talos using gcc-9.0 compiler (Dark red/blue), LLVM Clang-10 compiler (Light red/blue). $N \neq n$ for models with multi-DoF joints.} }
\label{pin_comparison}
\end{figure}

\begin{figure}[tb]
\hspace*{-0.5cm}
\includegraphics[width=9cm]{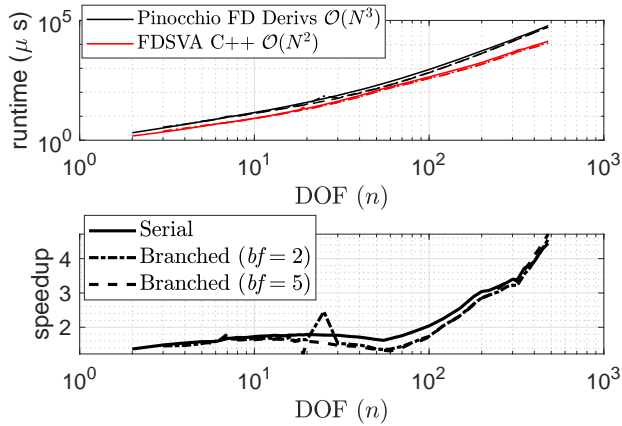}
\caption{a) CPU runtime comparison of \red{Pinocchio FD Derivs with FDSVA C++} for the first-order partial derivatives of FD for \red{serial and branched trees ($N=n$ for revolute joints).  Serial chain (bold line), Branched chain (\textit{bf}=2, dashed dotted line), Branched chain (\textit{bf}=5, dashed line)} b) Speedup plots show FDSVA C++ outperforms Pinocchio FD Derivs \red{for all N.}}
\label{fdsva_fdcr_pin}
\end{figure}

IDSVA in ground coordinates is implemented in C++ within the Pinochhio framework. This strategy enables direct comparison with Pinocchio's original ID partial derivatives~\cite{car_code}. \red{Fig.~\ref{SVA_vs_RNEA_FO_fig} shows the comparison of the two methods for serial and branched kinematic trees with some branching factor \textit{bf} ~\cite{rbd}}. \red{For a serial chain with $N=100$, a speedup of $2 \times$ is found using the gcc-9.0 compiler and turbo boost off.} \red{Results with implementation of IDSVA C++ to multi-DoF joint robots is shown in Fig.~\ref{pin_comparison} using the LLVM Clang-10 and gcc-9.0 compiler. For a 50 link floating-base humanoid Talos, IDSVA has a speedup of 1.4x over Pinocchio ID Derivs algorithm in Pinocchio \cite{car_code} using the gcc-9.0 compiler.
A fundamental similarity is found between the Pinocchio ID Derivs (open source, but unpublished) and IDSVA. Both the algorithms essentially calculate the same quantities, but a more efficient restructuring of the backward pass in IDSVA C++ led to the speedups shown in Fig.~\ref{pin_comparison}. In comparison to an $\mathcal{O}(d)$ innermost second backward pass over the ancestors of body $i$ in the Pinocchio ID Derivs code, IDSVA uses a matrix-matrix multiplication for the subtree of body $i$ on lines 19 and 20, leading to speed boosts. 
This small change makes the state-of-the-art performance of Pinocchio even better, and has been included in more recent releases. 
Fig.~\ref{fdsva_fdcr_pin} shows a comparison in the  CPU  runtimes for \red{FDSVA C++ (implemented in Pinocchio) with Pinocchio FD Derivs for serial and branched kinematic trees.}
 An open source Pinocchio implementation of IDSVA is available at \cite{cppsource} and a MATLAB version at \cite{matlabsource}.
 }
 
\subsection{Comparing CPU Runtime for AZA vs. DMM in C++ with Pinocchio Implementation Accompanying \cite{car}}
\begin{figure}[tb]
\includegraphics[width=8.5cm]{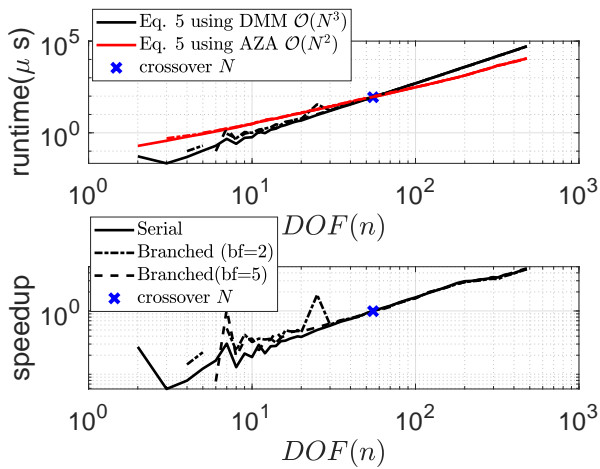}
\caption{a) \blue{Comparison of DMM vs.~AZA runtime implemented in Pinocchio C++ framework (using gcc-9.0) shows the cross-over $N$ at 50 for serial chain (solid), branched chain (\textit{bf}=2, dashed dotted), and branched chain (\textit{bf}=5, dashed). b) Speedup of AZA to DMM on a log-log scale grows linearly with $n$. $N=n$ for kinematic trees with revolute joints.}}
\label{dmm_vs_aza}
\end{figure}

AZA is implemented within the Pinocchio~\cite{car_code} framework. Pinocchio's $\M^{-1}$ algorithm~\cite{InverseMassMatrix} is modified to include the AZA. \blue{In Fig.~\ref{dmm_vs_aza}, the "crossover" $N$ denotes the point below which the DMM performs better than the AZA.}  This point depends on the hardware and compiler optimization settings used to implement the algorithm, but for high $N$, the $\mathcal{O}(N^2)$ AZA is efficient because it avoids the expensive matrix-matrix product. Table~\ref{table_ipr} gives the cross-over $N$ values (for kinematic trees) above which AZA performs better than DMM for different compiler settings. The effect of this cross-over $N$ can be seen in Fig.~\ref{fdsva_fdcr_pin} for the FDSVA curve at $N=50$, where the order of the algorithm changes from $\mathcal{O}(N^3)$ to $\mathcal{O}(N^2)$ due to switch from DMM to AZA to get Eq.~\ref{car_FO_eqn}.

\begin{table}[tb]
\centering
\begin{tabular}{ |c|c|c| } 
 \hline
 Compiler & AVX Off & AVX On \\ 
  \hline
 gcc-9.0 & 50 & 450 \\ 
 clang-10 & 80 & 350 \\ 
 \hline
\end{tabular}
 \caption{Cross-over $N$ (DoF) using different compilers and Autovectorization (\texttt{march=native}) settings.}
  \label{table_ipr}
\end{table}


\section{Implementation Considerations}
\label{sec:efficient_SVA}

\section{Conclusions}
In this work, we present \red{closed form partial derivatives of inverse dynamics, along with an efficient algorithm to calculate the inverse and forward dynamics partials for robots with multi-DoF joints and a floating base}. Several algorithmic optimizations and Spatial Vector Algebra (SVA) identities are exploited to enable an efficient implementation. \red{The method provides a $1.4 \times$ speedup for the TALOS humanoid model over the state-of-the-art Pinocchio FD derivatives using the gcc compiler and a $1.2 \times$ speedup using the Clang-10 compiler.} The reduction in runtime for partial derivatives enables faster optimization algorithms for both on-line and off-line applications. These improved timings can ultimately lead to better motion planning of legged and industrial robots.

\section{Acknowledgement}
This work was partially supported by the National Science Foundation EAGER grant CMMI-1835013. The authors thank Pinocchio developers Justin Carpentier \& Nicolas Mansard for important feedback and Wolfgang Merkt for benchmarking the IDSVA algorithm.


\appendix

\blue{
\subsection{Spatial Vector Algebra}
\label{sva_intro}
A body $k$ with spatial velocity ${}^{k}\v_{k} \in M^{6}$ in the body frame is decomposed in its angular and linear components as:

  \begin{align}
    {}^{k}\v_{k} &= \begin{bmatrix}
           {}^{k}\omega_{k} \\
           {}^{k}v_{k}
         \end{bmatrix}
  \end{align}
where ${}^{k}\omega_{k} \in \mathbb{R}^{3}$ is the angular velocity of the body in a coordinate frame fixed to the body, while ${}^{k}v_{k} \in \mathbb{R}^{3}$ is the linear velocity of the body-fixed point at the origin of the body frame. Spatial vectors can also be expressed in the ground frame. For example, the spatial velocity of the body $k$ in the ground frame is denoted as ${}^{0}\v_{k}$. In this case, the linear velocity is associated with the body-fixed point on body $k$ that is coincident with the origin of the ground frame. The net spatial force ${}^{0}\f_{k} \in F^{6}$ defined in Eq.~\ref{fvec_defn} on the body can be calculated from the spatial equation of motion (Eq.\ref{sp_eqn_of_motion}):

  \begin{align}
    {}^{0}\f_{k} &= \begin{bmatrix}
           {}^{0}n_{k} \\
           {}^{0}f_{k}
         \end{bmatrix}
         \label{fvec_defn}
  \end{align}
  
\begin{equation}
       {}^{0}\f_{k} = {}^{0}\I_{k}{}^{0}\a_{k} + {}^{0}\v_{k} \times^*{}^{0}\I_{k} {}^{0}\v_{k}
       \label{sp_eqn_of_motion}
\end{equation}

where ${}^{0}n_{k} \in \mathbb{R}^{3}$ is the net moment on the body about the origin of the ground frame, $ {}^{0}f_{k} \in \mathbb{R}^{3}$ is the net linear force on body, ${}^{0}\I_{k}$ is the spatial inertia of the body $k$ that maps motion vectors to force vectors, and ${}^{0}\a_{k} \in M^{6}$ is the spatial acceleration of the body. The transformation matrix ${}^{i}\boldsymbol{X}_{j}$ is used to transform vectors in frame $j$ to frame $i$ is defined as:

\begin{align}
    {}^{i}\boldsymbol{X}_{j} = 
    \begin{bmatrix}
        {}^{i}\boldsymbol{R}_{j} & \bf{0}  \\
       -{}^{i}\boldsymbol{R}_{j} (\p_{i/j} \times) &  {}^{i}\boldsymbol{R}_{j} 
    \end{bmatrix}
\end{align}
where ${}^{i}\boldsymbol{R}_{j} \in \mathbb{R}^{3 \times 3}$ is the rotation matrix from frame $j$ to frame $i$, $\p_{i/j} \in \mathbb{R}^{3}$ is the Cartesian vector from origin of frame $j$ to $i$, and $\bf{0}$ is the 3 $\times$ 3 zero matrix. $\p \times$ is the 3D vector cross product on the elements of $\p$, defined as:
\begin{align}
    \p \times =   \begin{bmatrix}
        0 & -p_{z}  & p_{y}\\
        p_{z} & 0  & -p_{x}\\
        -p_{y} & p_{x}  & 0
    \end{bmatrix}
\end{align}

\noindent The spatial transformation matrix ${}^{0}\boldsymbol{X}_{k}$ can be used to obtain spatial velocity vector ${}^{0}\v_{k}$ from the vector ${}^{k}\v_{k}$ as:

\begin{equation}
    {}^{0}\v_{k} = {}^{0}\boldsymbol{X}_{k} {}^{k}\v_{k}
\end{equation}

\noindent A spatial cross-product operator between two motion vectors ($\v$,$\u$), written as $(\v \times) \u$, is given by (Eq.~\ref{cross_defn}). This operation can be understood as providing the time rate of change of $\u$, when $\u$ is moving with a spatial velocity $\v$. A spatial cross-product between a motion and a force vector is written as $(\v \times^*) \f$, and defined in Eq.~\ref{cross_f_defn}. 
\begin{align}
    \v \times = \begin{bmatrix}
        \omega \times & \bf{0}  \\
       v \times &  \omega \times
    \end{bmatrix}
    \label{cross_defn}
\end{align}

\begin{align}
    \v \times^* = \begin{bmatrix}
        \omega \times & v \times  \\
        \bf{0} &  \omega \times
    \end{bmatrix}
    \label{cross_f_defn}
\end{align}

\noindent An operator $\crff$ (Eq.~\ref{crff_oper}) is  defined by swapping the order of the cross product, such that $(\f \crff) \v = (\v \times ^*) \f $~\cite{eche}. 

\begin{align}
    \f \crff = \begin{bmatrix}
        -n \times & -f \times  \\
        -f \times & \bf{0}
    \end{bmatrix}
    \label{crff_oper}
\end{align}

\noindent Hence, the three spatial vector cross-product operators defined above map between the motion and the force vector space as:~\cite{rbd}  

\begin{equation}
\begin{aligned}
&    \times : M^{6} \times M^{6} \xrightarrow[]{} M^{6} \\
&    \times^{*} : M^{6} \times F^{6} \xrightarrow[]{} F^{6} \\
&    \crff : F^{6} \times M^{6} \xrightarrow[]{} F^{6}
\end{aligned}
\end{equation}

\subsection{Properties of Spatial Vectors}
\label{props}
Assuming $\u,\v,\m,\v_{1},\v_{2} \in M^{6}$, and $\f \in F^{6}$, many spatial vector properties~\cite{rbd} are utilized herein:

\begin{enumerate}[label=P{{\arabic*}}.]
    \item $(\v \times \m) \times = (\v \times )( \m \times) - (\m \times)(\v \times)$ 
    \item $(\v \times \m) \times^{*} = (\v \times^{*} )( \m \times^*) - (\m \times^*)(\v \times^*)$ 
    \item $(\v \times^{*} \f) \crff = (\v \times^{*} )(\f \crff) - (\f \crff)(\v \times)$ 
    \item $(\v_{1} \times \v_2)^{T}\f = -\v_2^T (\v_1 \times^* \f)  $
    \item $(\v_1 \times^* \f)^T \v_2 = -\f^T(\v_1 \times \v_2)$
    \item $(\u \times \v)^T  = -\v^{T}(\u \times^*) $
    \item $(\u \times^* \f) ^T = -\f ^T \u \times$
\end{enumerate}

\subsection{Multi-DoF joint Identities and Expressions}
\label{multi_dof_iden}
Identities are shown below for partial derivatives of common spatial quantities by perturbing a multi DoF joint position variable ($\q_{j}$). These identities are used to then derive the partial derivatives of inverse dynamics for a rigid-body system with multi-DoF joints and a floating base.

\begin{enumerate}[label=J{{\arabic*}}.]

         \item $\frac{\partial \Phibar_{i}}{\partial q_{j,p}}=
        \begin{cases}
          \phibar_{j,p} \times \Phibar_{i} , &~~~~~~~~~~~~~~~~~~~~ \text{if}\ j \preceq i \\
          0, &~~~~~~~~~~~~~~~~~~~~ \text{otherwise}
        \end{cases}$

       \item $ \frac{\partial (\v_{i} \times^{*} \f)}{\partial \q_{j}}=
        \begin{cases}
         \f \crff \Big( (\v_{\lambda(j)} - \v_{i})  \times \Phibar_{j} \Big) ,  & \text{if}\ j \preceq i \\
          0,  & \text{otherwise}
        \end{cases}$
   
      \item $ \frac{\partial (\Phibar_{i}^{T} \f)}{\partial \q_{j}}=
    \begin{cases}
      -\Phibar_{i}^{T} (f \crff  \Phibar_{j}) , &~~~~~~~~~~~~ \text{if}\ j \preceq i \\
      0, &~~~~~~~~~~~~ \text{otherwise}
    \end{cases}$

     \item $\frac{\partial (\I_{i} \a)}{\partial \q_{j}} =
    \begin{cases}
     (\I_{i} \a) \crff \Phibar_{j} +\I_{i}(\a \times \Phibar_{j})   , & \text{if}\ j \preceq i \\
      0, & \text{otherwise}
    \end{cases}$

     \item $\frac{\partial (\I_{i} \v_{i})}{\partial \q_{j}}  =
    \begin{cases}
    (\I_{i} \v_{i}) \crff \Phibar_{j} + \I_{i}(\Psibardot_{j})  , &~~~ \text{if}\ j \preceq i \\
      0, &~~~ \text{otherwise}
    \end{cases}$

     \item $\frac{\partial \xibar_{i}}{\partial \q_{j}}=
    \begin{cases}
       (\v_{\lambda(j)} - \v_{i} )  \times  \Psibardot_{j} + \\
       ~~( \xibar_{\lambda(j)} - \xibar_{i}  ) \times  \Phibar_{j}   , &~~~~~~~~~~ \text{if}\ j \preceq i \\
      0, &~~~~~~~~~~ \text{otherwise}
    \end{cases}$

       \item $\frac{\partial \gammabar_{i}}{\partial \q_{j}}=
    \begin{cases}
       (\gammabar_{\lambda(j)} - \gammabar_{i})  \times \Phibar_{j} , &~~~~~~~~~~~~ \text{if}\ j \preceq i \\
      0, &~~~~~~~~~~~~ \text{otherwise}
    \end{cases}$
    
        \item $\frac{\partial \v_{i}}{\partial \dot{\q}_{j}}=
    \begin{cases}
          \Phibar_{j}, &~~~~~~~~~~~~~~~~~~~~~~~~~~~~~~ \text{if}\ j \preceq i \\
      0, &~~~~~~~~~~~~~~~~~~~~~~~~~~~~~~ \text{otherwise}
    \end{cases}$  
    
     \item $\frac{\partial \xibar_{i}}{\partial \dot{\q}_{j}}=
    \begin{cases}
    \Psibardot_{j} + \Phibardot_{j}-\v_{i}  \times \Phibar_{j}, &~~~~~~~~~~ \text{if}\ j \preceq i \\
      0, &~~~~~~~~~~ \text{otherwise}
    \end{cases}$
    
\end{enumerate}

 Identities listed above are derived in detail as follows:

\begin{enumerate}
 \item [J1.] For directional derivative of $\Phibar_{i}$ along the $p^{th}$ free-mode of the joint $j$, total derivative with respect to time is taken for the numerator and denominator as:

\begin{equation}
    \frac{\partial \Phibar_{i}}{\partial q_{j,p}}=\frac{\partial{\dot{\Phibar}}_{i}}{\partial \dot{q}_{j,p}}
\end{equation}

Using the definition of $\Phibardot_{i}$ as $\Phibardot_{i}=\v_{i} \times \Phibar_{i}$, and the definition of $\v_{i}$:

\begin{equation}
    \frac{\partial \Phibar_{i}}{\partial q_{j,p}}=\frac{\partial (\sum_{l \preceq i} \Phibar_{l} \qd_{l} \times \Phibar_{i} )}{\partial \dot{q}_{j,p}}
\end{equation}

Only a single term remains for the case $j \preceq i$:

\begin{equation}
    \frac{\partial \Phibar_{i}}{\partial q_{j,p}}= \phibar_{j,p}\times \Phibar_{i} 
    \label{J1_eqn}
\end{equation}   

 \item[J2.] The directional derivative of the spatial velocity of a body $i$ along the $p^{th}$ free-mode of the joint $j$, where $j \preceq i$ is given as:

\begin{equation}
    \frac{\partial \v_{i}}{\partial q_{j,p}} = \sum_{l \preceq i} \frac{\partial \Phibar_{l}}{\partial q_{j,p}}  \qd_{l}
\end{equation}
Using J1, we get:

\begin{equation}
    \frac{\partial \v_{i}}{\partial q_{j,p}} = \sum_{j \preceq l \preceq i} \phibar_{j,p} \times \Phibar_{l} \qd_{l}
\label{dv_eq2}
\end{equation}

By switching signs, Eq.~\ref{dv_eq2} can also be written as:

\begin{equation}
    \frac{\partial \v_{i}}{\partial q_{j,p}} = - \sum_{j \preceq l \preceq i} \Phibar_{l} \qd_{l}  \times \phibar_{j,p}
\label{dv_eq3}
\end{equation}

Eq~\ref{dv_eq3} can be written for each $p^{th}$ mode of the joint $j$ to get the partial derivative (of size $6 \times n_{j}$, where $n_j$ is the number of DoF for joint $j$) of $\v_{i}$ with respect to $\q_{j}$  as:

\begin{equation}
    \frac{\partial \v_{i}}{\partial \q_{j}} = - \sum_{j \preceq l \preceq i} \Phibar_{l} \qd_{l}  \times \Phibar_{j}
\label{dv_eq4}
\end{equation}
Using the definition of $\v_{l}$, Eq.~\ref{dv_eq4} now is: 

\begin{equation}
    \frac{\partial \v_{i}}{\partial \q_{j}} = (\v_{\lambda(j)} - \v_{i})  \times \Phibar_{j}
\label{dvi_dqj_iden}
\end{equation}

Using Eq.~\ref{dvi_dqj_iden}, the partial derivative of $\v_{i} \times \a$ with respect to $\q_j$ for $j \preceq i$, where $\a \in M^{6}$ is any fixed motion vector is given as:

\begin{equation}
    \frac{\partial (\v_{i} \times \a)}{\partial \q_{j}} = -\a \times \Big( (\v_{\lambda(j)} - \v_{i})  \times \Phibar_{j}  \Big)
\label{dva_eq1}
\end{equation}

Similar to Eq.~\ref{dva_eq1}, the partial derivative of $\v_{i} \times^{*} \f$, where $\f \in F^{6}$ is any fixed force vector, is calculated as:

\begin{equation}
   \frac{\partial (\v_{i} \times^{*} \f)}{\partial \q_{j}} = \f \crff \Big( (\v_{\lambda(j)} - \v_{i})  \times \Phibar_{j}  \Big)
\label{dvcrossstara_eq1}   
\end{equation}

\item[J3.] For any fixed force vector $\f$, the partial derivative of $\Phibar_{i}^{T} \f$ with respect to $\q_j$ for $j \preceq i$ is calculated. Using J1:

\begin{equation}
 \frac{\partial (\Phibar_{i}^{T} \f)}{\partial q_{j,p}} =  (\phibar_{j,p} \times \Phibar_{i})^{T} \f 
 \label{dphif_eq1}
\end{equation}

\begin{equation}
 \frac{\partial (\Phibar_{i}^{T} \f)}{\partial q_{j,p}} = - \Phibar_{i}^{T} (\phibar_{j,p} \times^{*} ) \f 
 \label{dphif_eq2}
\end{equation}

Eq.~\ref{dphif_eq2} can be written for each DoF of the joint $j$. Hence, the partial derivative with respect to $\q_{j}$ is:

\begin{equation}
 \frac{\partial (\Phibar_{i}^{T} \f)}{\partial \q_{j}} = - \Phibar_{i}^{T} (\f \crff  \Phibar_{j}) 
 \label{dphif_eq3}
\end{equation}

\item[J4.] For any fixed motion vector $\a$, the partial derivative of $\I_{i} \a$ is calculated with respect to $\q_{j}$ for $j \preceq i$. The directional derivative of $\I_{i}$ in the direction of $p^{th}$ free-mode of joint $j$~\cite{rbd} is:

\begin{equation}
    \frac{\partial \I_{i}}{\partial q_{j,p}}=   \phibar_{j,p} \times^* \I_{i}- \I_{i}(\phibar_{j,p} \times)
    \label{I5_eqn}
\end{equation}

Multiplying $\a$ on both sides result in:

\begin{equation}
    \frac{\partial (\I_{i} \a)}{\partial q_{j,p}} =  \phibar_{j,p} \times^* \I_{i} \a- \I_{i}(\phibar_{j,p} \times \a)   
    \label{dIa_eq1}
\end{equation}

Similar to Eq~\ref{dphif_eq2}, Eq~\ref{dIa_eq1} can be written for each DoF of joint $j$ collectively as:

\begin{equation}
    \frac{\partial (\I_{i} \a)}{\partial \q_{j}} =  (\I_{i} \a) \crff \Phibar_{j} + \I_{i}(\a \times \Phibar_{j})   
    \label{dIa_eq2}
\end{equation}

\item[J5.] The partial derivative of $\I_{i} \v_{i}$ with respect to $\q_{j}$ for $j \preceq i$ is now calculated using the identities derived above. Using the product rule:

\begin{equation}
    \frac{\partial (\I_{i} \v_{i})}{\partial \q_{j}} = \frac{\partial (\I_{i} )}{\partial \q_{j}} \v_{i} + \I_{i} \frac{\partial (\v_{i} )}{\partial \q_{j}}     
    \label{dIv_eq1}
\end{equation}

For the first term in Eq.~\ref{dIv_eq1} on the RHS, $\v_{i}$ is assumed to be a fixed motion vector. Hence, the identity J4 can be used. For the second term in Eq.~\ref{dIv_eq1}, Eq.~\ref{dvi_dqj_iden} is used as:

\begin{equation}
    \begin{aligned}
    \frac{\partial (\I_{i} \v_{i})}{\partial \q_{j}} = (\I_{i} \v_{i}) \crff \Phibar_{j} + \I_{i}(\v_{i} \times \Phibar_{j})  +  \\
     \I_{i} ( (\v_{\lambda(j)} - \v_{i} )  \times \Phibar_{j})     
    \label{dIv_eq2}
    \end{aligned}
\end{equation}

Expanding terms:

\begin{equation}
    \begin{aligned}
    \frac{\partial (\I_{i} \v_{i})}{\partial \q_{j}} = (\I_{i} \v_{i}) \crff \Phibar_{j} + \I_{i}(\v_{i} \times \Phibar_{j})  + \\
    \I_{i} ( \v_{\lambda(j)} \times \Phibar_{j})  - \I_{i} (\v_{i} \times \Phibar_{j}) 
    \label{dIv_eq3}
    \end{aligned}
\end{equation}

Upon cancellations and simplification, Eq.~\ref{dIv_eq3} becomes:

\begin{equation}
    \frac{\partial (\I_{i} \v_{i})}{\partial \q_{j}} = (\I_{i} \v_{i}) \crff \Phibar_{j} + \I_{i}(\Psibardot_{j})
    \label{dIv_eq4}
\end{equation}

where $\Psibardot_{j}$ is defined in Eq.~\ref{psidot_psidotdot_defn}.

\item[J6.] Using identities derived above, and the definition of $\xibar_{i}$, the partial derivative of $\xibar_{i}$ with respect to $\q_{j}$ for $j \preceq i$ is calculated as: 
\begin{equation}
    \frac{\partial \xibar_{i}}{\partial \q_{j}} = \sum_{l \preceq i} \frac{\partial (\v_{l} \times \vJ{l})}{\partial \q_{j}}  
    \label{dxi_eq1}
\end{equation}
where $\vJ{l}$ is the joint velocity defined as:

\begin{equation}
    \vJ{l} = \Phibar_{l} \qd_{l}
    \label{vj_defn}
\end{equation}

 Using the product rule of differentiation:

\begin{equation}
    \frac{\partial \xibar_{i}}{\partial \q_{j}} = \sum_{j \preceq l \preceq i} \frac{\partial (\v_{l} \times )}{\partial \q_{j}}  \vJ{l} + \v_{l} \times \frac{\partial (\vJ{l} )}{\partial \q_{j}}  
    \label{dxi_eq2}
\end{equation}

The partial derivative of the joint velocity $\vJ{i}$ with respect to $\q_{j}$ for $j \preceq i$ is calculated. Using the definition of $\vJ{i}$ (Eq.~\ref{vj_defn}) we get:

\begin{equation}
    \vJ{i} = \Phibar_{i} \qd_{i}
    \label{vj_defn}
\end{equation}

\begin{equation}
    \frac{\partial \vJ{i}}{\partial q_{j,p}} = \frac{\partial (\Phibar_{i} \qd_{i})}{\partial q_{j,p}}  
    \label{dvj_eq1}
\end{equation}

Using identity J1, we get

\begin{equation}
    \frac{\partial \vJ{i}}{\partial q_{j,p}} = \phibar_{j,p} \times \Phibar_{i} \qd_{i}
    \label{dvj_eq2}
\end{equation}

Simplifying Eq.~\ref{dvj_eq2},

\begin{equation}
    \frac{\partial \vJ{i}}{\partial q_{j,p}} = - \vJ{i} \times \phibar_{j,p}
    \label{dvj_eq3}
\end{equation}

Collectively, for all DoF of joint $j$, Eq.~\ref{dvj_eq3} can be written as:

\begin{equation}
    \frac{\partial \vJ{i}}{\partial \q_{j}} = - \vJ{i} \times \Phibar_{j}
    \label{dvji_dqj}
\end{equation}

Treating $\vJ{l}$ as a fixed motion vector in the first term in Eq.~\ref{dxi_eq2} on the RHS, Eq.~\ref{dvi_dqj_iden} is used.  Eq.~\ref{dva_eq1} is used for the second term.

\begin{equation}
    \begin{aligned}
        \frac{\partial \xibar_{i}}{\partial \q_{j}} = \sum_{j \preceq l \preceq i} -(\vJ{l}) \times \big( (\v_{\lambda(j)} - \v_{l} )  \times \Phibar_{j} \big ) \\
        + \v_{l} \times (-\vJ{l} \times \Phibar_{j})
        \label{dxi_eq3}
    \end{aligned}
\end{equation}

Expanding terms, and using the definition of $\Psibardot_{j}$ (Eq.~\ref{psidot_psidotdot_defn}):

\begin{equation}
    \begin{aligned}
       \frac{\partial \xibar_{i}}{\partial \q_{j}} = \sum_{j \preceq l \preceq i} -(\vJ{l} \times \Psibardot_{j}) + \vJ{l} \times (\v_{l} \times \Phibar_{j})  \\
       - \v_{l} \times (\vJ{l} \times \Phibar_{j})
        \label{dxi_eq4}
    \end{aligned}
\end{equation}

Combining terms and simplifying,

\begin{equation}
    \begin{aligned}
        \frac{\partial \xibar_{i}}{\partial \q_{j}} = \sum_{j \preceq l \preceq i} -(\vJ{l} \times \Psibardot_{j}) + ( \vJ{l} \times \v_{l} \times - \\
        \v_{l} \times \vJ{l} \times) \Phibar_{j}   \label{dxi_eq5}
    \end{aligned}
\end{equation}

Using spatial vector cross property P1, definition of $\xibar_{i}$ and simplifying,

\begin{equation}
    \begin{aligned}
        \frac{\partial \xibar_{i}}{\partial \q_{j}} = \sum_{j \preceq l \preceq i} -(\vJ{l} \times \Psibardot_{j}) - ( \v_{l} \times \vJ{l} ) \times  \Phibar_{j}  
       \label{dxi_eq6}
    \end{aligned}
\end{equation}

Summing over the terms and simplifying, we get:

\begin{equation}
    \begin{aligned}
       \frac{\partial \xibar_{i}}{\partial \q_{j}} =   (\v_{\lambda(j)} - \v_{i})  \times  \Psibardot_{j} + 
       ( \xibar_{\lambda(j)} - \xibar_{i}) \times  \Phibar_{j}  
       \label{dxi_eq7}
    \end{aligned}
\end{equation}

\item[J7.] Using the definition of $\gammabar_{i}$, the directional derivative of $\gammabar_{i}$ along the $p^{th}$ free-mode of the joint $j$ for $j \preceq i$ is calculated as:

\begin{equation}
    \frac{\partial \gammabar_{i}}{\partial q_{j,p}} = \sum_{l \preceq i} \frac{\partial \Phibar_{l}}{\partial q_{j,p}}  \qdd_{l}
    \label{dgamma_eq1}
\end{equation}
Using J1, we get:

\begin{equation}
    \frac{\partial \gammabar_{i}}{\partial q_{j,p}} = \sum_{j \preceq l \preceq i} \phibar_{j,p} \times \Phibar_{l} \qdd_{l}
\label{dgamma_eq2}
\end{equation}

Eq.~\ref{dgamma_eq2} can also be written as

\begin{equation}
    \frac{\partial \gammabar_{i}}{\partial q_{j,p}} = - \sum_{j \preceq l \preceq i} \Phibar_{l} \qdd_{l}  \times \phibar_{j,p}
\label{dgamma_eq3}
\end{equation}

Collectively for all DoF of joint $j$ , Eq.~\ref{dgamma_eq3} is written as:

\begin{equation}
    \frac{\partial \gammabar_{i}}{\partial \q_{j}} =  - \sum_{j \preceq l \preceq i} \Phibar_{l} \qdd_{l}  \times \Phibar_{j}
\label{dgamma_eq4}
\end{equation}

Using the definition of $\gammabar_{l}$, Eq.~\ref{dgamma_eq4} becomes: 

\begin{equation}
    \frac{\partial \gammabar_{i}}{\partial \q_{j}} = (\gammabar_{\lambda(j)} - \gammabar_{i})  \times \Phibar_{j}
\label{dgamma_eq5}
\end{equation}

\item[J8.] The partial derivative of $\v_{i}$ is calculated with respect to $\dot{\q}_{j}$ for $j \preceq i$. Using the definition of $\v_{i}$, we get:

\begin{equation}
    \frac{\partial \v_{i}}{\partial \dot{\q}_{j}} = \sum_{j \preceq l \preceq i} \Phibar_{l} \frac{\partial  \qd_{l} }{\partial \dot{\q}_{j}} 
    \label{dv_dqdot_eq1}
\end{equation}

Summing over all indices $l$, all the terms vanish except the ones pertaining to the index $j$ resulting in:

\begin{equation}
    \frac{\partial \v_{i}}{\partial \dot{\q}_{j}} = \Phibar_{j}
    \label{dv_dqdot_eq2}
\end{equation}

\item[J9.] Using the definition of $\xibar_{i}$, the partial derivatives of $\xibar_{i}$ with respect to $\dot{\q}_{j}$ for $j \preceq i$ is:

\begin{equation}
    \frac{\partial \xibar_{i}}{\partial \dot{\q}_{j}} = \sum_{l \preceq i} \frac{\partial (\v_{l} \times \Phibar_{l} \dot{\q}_{l})}{\partial \dot{\q}_{j}}  
    \label{dxi_qdot_eq1}
\end{equation}

Using the product rule of differentiation, we get:

\begin{equation}
    \frac{\partial \xibar_{i}}{\partial \dot{\q}_{j}} = \sum_{j \preceq l \preceq i} \frac{\partial (\v_{l} \times )}{\partial \dot{\q}_{j}}  \Phibar_{l} \dot{\q}_{l} + \v_{l} \times \frac{\partial (\Phibar_{l} \dot{\q}_{l} )}{\partial \dot{\q}_{j}}  
    \label{dxi_qdot_eq2}
\end{equation}

Using J8, summing over the index $l$ results in:

\begin{equation}
    \frac{\partial \xibar_{i}}{\partial \dot{\q}_{j}} =  (\v_{\lambda(j)} - \v_{i})  \times \Phibar_{j} + \Phibardot_{j}
    \label{dxi_qdot_eq3}
\end{equation}

Upon simplifying and using the definition of $\Psibardot_j$ (Eq.~\ref{psidot_psidotdot_defn}), we get:

\begin{equation}
    \frac{\partial \xibar_{i}}{\partial \dot{\q}_{j}} =  \Psibardot_{j}+ \Phibardot_{j} -\v_{i}\times \Phibar_{j} 
    \label{dxi_qdot_eq4}
\end{equation}

\end{enumerate}

\subsection{Partial Derivatives of $ID$ w.r.t $\q$: Derivations}
\label{partials_details}
{\noindent \bf Partial Derivatives of $[ \M(\q) \qdd]_{i}$:}
\begin{enumerate}
  \item Case when $j \preceq i$ \\
  Using product rule of differentiation in Eq.~\ref{mqdotdot_exp_2}, we get:
  \begin{equation}
      \begin{aligned}
        \frac{\partial [ \M(\q) \qdd]_{i}}{\partial \q_{j}} =  \frac{\partial (\Phibar_{i}^{T})}{\partial \q_{j}} \sum_{k \succeq i} [\I_{k} \gammabar_{k}]   +  \\ 
        \Phibar_{i}^{T}  \sum_{k \succeq i} [\frac{\partial \I_{k}}{\partial \q_{j}} \gammabar_{k} + \I_{k} \frac{\partial \gammabar_{k}}{\partial \q_{j}} ]  
      \end{aligned}
\end{equation}
 Using identities J3, J4, and J7, we get:
  
   \begin{equation}
      \begin{aligned}
         \frac{\partial [ \M(\q) \qdd]_{i}}{\partial \q_{j}} =  -\Phibar_{i}^{T} \Big(\sum_{k \succeq i} [\I_{k} \gammabar_{k}]  \crff \Big) \Phibar_{j}  + \\
         \Phibar_{i}^{T}  \sum_{k \succeq i} \Big[(\I_{k} \gammabar_{k}) \crff \Phibar_{j} +  \I_{k}(\gammabar_{k} \times \Phibar_{j})+\\ 
         \I_{k}  (\gammabar_{\lambda(j)} - \gammabar_{k} )  \times \Phibar_{j} \Big]  
      \end{aligned}
\end{equation} 
  
\noindent Upon cancellations and simplifications, the final expression is:
 
   \begin{equation}
         \frac{\partial [ \M(\q) \qdd]_{i}}{\partial \q_{j}} =  \Phibar_{i}^{T} [  \I_{i}^{C}  \gammabar_{\lambda(j)}   \times \Phibar_{j} ]  
        \label{partial_Mqddot_3}
\end{equation}

  \item Case when $j \succ i$ \\
  For this case, identities J3, J4 and J7 are used as:
  
  \begin{equation}
         \frac{\partial [ \M(\q) \qdd]_{i}}{\partial \q_{j}} =   \Phibar_{i}^{T}  \sum_{k \succeq i} [\frac{\partial \I_{k}}{\partial \q_{j}} \gammabar_{k} + \I_{k} \frac{\partial \gammabar_{k}}{\partial \q_{j}} ]  
        \label{partial_Mqddot_4}
\end{equation}
 
Expanding:
  
   \begin{equation}
      \begin{aligned}
       \frac{\partial [ \M(\q) \qdd]_{i}}{\partial \q_{j}} = 
     \Phibar_{i}^{T}  \sum_{ k \succeq j} \Big[(\I_{k} \gammabar_{k}) \crff \Phibar_{j} +  \\
     \I_{k}(\gammabar_{k} \times \Phibar_{j})+   \I_{k}  (\gammabar_{\lambda(j)} - \gammabar_{k} )  \times \Phibar_{j} \Big]  
        \label{partial_Mqddot_5}
      \end{aligned}
\end{equation} 

Upon cancellations and simplification, we get the following expression.

\begin{equation}
\frac{\partial [ \M(\q) \qdd]_{i}}{\partial \q_{j}} = \Phibar_{i}^{T} [\etabar_{j}^{C} \crff \Phibar_{j} + \I_{j}^{C} \gammabar_{\lambda(j)}\times \Phibar_{j}  ]
    \label{partial_Mqddot_6}
\end{equation}

 \end{enumerate}

{\noindent \bf 
Partial Derivative of $[\C(\q,\qd) \qd]_{i}$:}

\begin{enumerate}
 \item Case when $j \preceq i$ \\
 
  Using product rule of differentiation in Eq.~\ref{cqdot_exp_2}, we get:
  \begin{equation}
      \begin{aligned}
        \frac{\partial  [\,\C \qd\,]_{i}}{\partial \q_{j}} =  \frac{\partial (\Phibar_{i}^{T})}{\partial \q_{j}} \sum_{k \succeq i} [\v_{k} \times^* \I_{k} \v_{k}+ \I_{k}  \xibar_{k}] +\\ 
        \Phibar_{i}^{T}  \sum_{k \succeq i} \Big[\frac{\partial (\v_{k} \times^*)}{\partial \q_{j}} \I_{k} \v_{k} +  (\v_{k} \times^*) \frac{\partial (\I_{k} \v_{k})}{\partial \q_{j}} + \\
         \frac{\partial \I_{k} }{\partial \q_{j}} \xibar_{k} + \I_{k} \frac{\partial \xibar_{k} }{\partial \q_{j}} \Big]  
      \end{aligned}
\end{equation}
Using identities J2-J6, we get:
    
   \begin{equation}
      \begin{aligned}
         \frac{\partial  [\,\C \qd\,]_{i}}{\partial \q_{j}} = 
         -\Phibar_{i}^{T} \Big(\sum_{k \succeq i} [v_{k} \times^* \I_{k} \v_{k} +    \I_{k}  \xibar_{k}]  \crff \Big) \Phibar_{j} \\ + \Phibar_{i}^{T}  \sum_{k \succeq i} \Big[  (\I_{k} \v_{k}) \crff     \big( (\v_{\lambda(j)} - \v_{k} )   \times \Phibar_{j} \big ) + \\
         (\v_{k} \times^*) (\I_{k} \v_{k}) \crff \Phibar_{j} +  (\v_{k} \times^*) \I_{k}(\Psibardot_{j})  +   \\
         (\I_{k} \xibar_{k} ) \crff \Phibar_{j} +\I_{k}(\xibar_{k}  \times \Phibar_{j})  + \\
         \I_{k}( (\v_{\lambda(j)} - \v_{k} )  \times  \Psibardot_{j} +\\  ( \xibar_{\lambda(j)} - \xibar_{k} ) \times  \Phibar_{j}) \Big]  
      \end{aligned}
\end{equation}

Expanding terms, and using the property P3:

   \begin{equation}
      \begin{aligned}
         \frac{\partial  [\,\C \qd\,]_{i}}{\partial \q_{j}} = -\Phibar_{i}^{T} \Big(\sum_{k \succeq i} [\v_{k} \times^* (\I_{k} \v_{k}) \crff \Phibar_{j} - \\
         (\I_{k} \v_{k}) \crff \v_{k}\times \Phibar_{j}  +    (\I_{k}  \xibar_{k})\crff \Phibar_{j} ]   \Big)  +   \\ 
         \Phibar_{i}^{T}  \sum_{k \succeq i} \Big[  (\I_{k} \v_{k}) \crff \Psibardot_{j} -   (\I_{k} \v_{k}) \crff \v_{k}\times \Phibar_{j} + \\
         (\v_{k} \times^*) (\I_{k} \v_{k}) \crff \Phibar_{j} + 
         (\v_{k} \times^*) \I_{k}(\Psibardot_{j})  +   \\  (\I_{k} \xibar_{k} ) \crff \Phibar_{j} +\I_{k}(\xibar_{k}  \times \Phibar_{j})  + 
         \I_{k} \v_{\lambda(j)} \times \Psibardot_{j} - \\ \I_{k}( \v_{k} \times \Psibardot_{j})  +
         \I_{k}\xibar_{\lambda(j)} \times \Phibar_{j} - \\
         \I_{k} (\xibar_{k} \times \Phibar_{j}) \Big]  
      \end{aligned}
\end{equation}

Simplifying terms,

   \begin{equation}
      \begin{aligned}
         \frac{\partial  [\,\C \qd\,]_{i}}{\partial \q_{j}} = \Phibar_{i}^{T} \Big(\sum_{k \succeq i} [-\v_{k} \times^* (\I_{k} \v_{k}) \crff \Phibar_{j} +\\
         (\I_{k} \v_{k}) \crff \v_{k}\times \Phibar_{j}  -  
         (\I_{k}  \xibar_{k})\crff \Phibar_{j} ]   \Big)  +   \\     \Phibar_{i}^{T}  \sum_{k \succeq i} \Big[  (\I_{k} \v_{k}) \crff \Psibardot_{j} -  (\I_{k} \v_{k}) \crff \v_{k}\times \Phibar_{j}  + \\
         (\v_{k} \times^*) (\I_{k} \v_{k}) \crff \Phibar_{j} +  
         (\v_{k} \times^*) \I_{k}(\Psibardot_{j})  +  \\
         (\I_{k} \xibar_{k} ) \crff \Phibar_{j} + 
         \I_{k}(\xibar_{k}  \times \Phibar_{j})  +  \\
         \I_{k} \v_{\lambda(j)} \times \Psibardot_{j} 
         -  \I_{k}( \v_{k} \times \Psibardot_{j})  +\\
         \I_{k}\xibar_{\lambda(j)} \times \Phibar_{j} -
         \I_{k} (\xibar_{k} \times \Phibar_{j}) \Big]  
        \label{partial_Cqdot_2}
      \end{aligned}
\end{equation}

Cancelling terms,

   \begin{equation}
      \begin{aligned}
         \frac{\partial  [\,\C \qd\,]_{i}}{\partial \q_{j}} =     \Phibar_{i}^{T}  \sum_{k \succeq i} \Big[  (\I_{k} \v_{k}) \crff \Psibardot_{j}     +\\
         (\v_{k} \times^*) \I_{k}(\Psibardot_{j}) + 
         \I_{k} \v_{\lambda(j)} \times \Psibardot_{j} - \\
         \I_{k}( \v_{k} \times \Psibardot_{j})  + 
           \I_{k}\xibar_{\lambda(j)} \times \Phibar_{j}  \Big] 
        \label{partial_Cqdot_3}
      \end{aligned}
\end{equation}

Combining terms,

   \begin{equation}
      \begin{aligned}
         \frac{\partial  [\,\C \qd\,]_{i}}{\partial \q_{j}} =     \Phibar_{i}^{T}  \sum_{k \succeq i} \Big[  (\I_{k} \v_{k}) \crff \Psibardot_{j}    + \\
           (\v_{k} \times^*) \I_{k}(\Psibardot_{j})  +
           \I_{k} \v_{\lambda(j)} \times \Psibardot_{j} - \\
           \I_{k}( \v_{k} \times \Psibardot_{j})  +  \I_{k}\xibar_{\lambda(j)} \times \Phibar_{j}  \Big]  
        \label{partial_Cqdot_4}
      \end{aligned}
\end{equation}

Using the definition of $\B_{i}$ (Eq.~\ref{bl_term_defn}), and summing over the index $k$, we get:

   \begin{equation}
      \begin{aligned}
         \frac{\partial  [\,\C \qd\,]_{i}}{\partial \q_{j}} =     \Phibar_{i}^{T}  \Big[  2 \B_{i}^{C} \Psibardot_{j}  +\I_{i}^{C} \v_{\lambda(j)} \times \Psibardot_{j}   + \\
         \I_{i}^{C} \xibar_{\lambda(j)} \times \Phibar_{j}  \Big]  
        \label{partial_Cqdot_5}
      \end{aligned}
\end{equation}

 \item Case when $j \succ i $
 
 For this case, first the product rule of differentiation is used:
 
   \begin{equation}
      \begin{aligned}
        \frac{\partial  [\,\C \qd\,]_{i}}{\partial \q_{j}} =   \Phibar_{i}^{T}  \sum_{k \succeq i} \Big[\frac{\partial (\v_{k} \times^*)}{\partial \q_{j}} \I_{k} \v_{k} +\\
        (\v_{k} \times^*) \frac{\partial (\I_{k} \v_{k})}{\partial \q_{j}} +   \frac{\partial \I_{k} }{\partial \q_{j}} \xibar_{k} + \I_{k} \frac{\partial \xibar_{k} }{\partial \q_{j}} \Big]  
        \label{partial_Cqdot_6}
      \end{aligned}
\end{equation}
 
 Using identities J2-J6, we get:

   \begin{equation}
      \begin{aligned}
         \frac{\partial  [\,\C \qd\,]_{i}}{\partial \q_{j}} &=        \Phibar_{i}^{T}  \sum_{k \succeq j} \Big[  (\I_{k} \v_{k}) \crff  \big( (\v_{\lambda(j)} - \v_{k} )  \times \Phibar_{j} \big )  \\&
         +(\v_{k} \times^*) (\I_{k} \v_{k}) \crff \Phibar_{j} + 
         (\v_{k} \times^*) \I_{k}(\Psibardot_{j})  + \\&
         (\I_{k} \xibar_{k} ) \crff \Phibar_{j} + 
         \I_{k}(\xibar_{k}  \times \Phibar_{j})  +  \\&
         \I_{k}( (\v_{\lambda(j)} - \v_{k} )  \times  \Psibardot_{j} + 
         ( \xibar_{\lambda(j)} - \xibar_{k} ) \times  \Phibar_{j}) \Big]  
        \label{partial_Cqdot_7}
      \end{aligned}
\end{equation}

 Expanding terms 

   \begin{equation}
      \begin{aligned}
         \frac{\partial  [\,\C \qd\,]_{i}}{\partial \q_{j}} &=   \Phibar_{i}^{T}  \sum_{k \succeq j} \Big[  (\I_{k} \v_{k}) \crff \Psibardot_{j} - \\&
         (\I_{k} \v_{k}) \crff \v_{k}\times \Phibar_{j}  +
         (\v_{k} \times^*) (\I_{k} \v_{k}) \crff \Phibar_{j} + \\&
         (\v_{k} \times^*) \I_{k}(\Psibardot_{j})  +  (\I_{k} \xibar_{k} ) \crff \Phibar_{j} +\\& 
         \I_{k}(\xibar_{k}  \times \Phibar_{j})  + 
         \I_{k} \v_{\lambda(j)} \times \Psibardot_{j} -  \\& 
         \I_{k}( \v_{k} \times \Psibardot_{j})+
         \I_{k}\xibar_{\lambda(j)} \times \Phibar_{j} - \\&
         \I_{k} (\xibar_{k} \times \Phibar_{j}) \Big]  
        \label{partial_Cqdot_8}
      \end{aligned}
\end{equation}

Cancelling, re-arranging terms,

   \begin{equation}
      \begin{aligned}
         \frac{\partial  [\,\C \qd\,]_{i}}{\partial \q_{j}} &=   \Phibar_{i}^{T}  \sum_{k \succeq j} \Big[  (\I_{k} \v_{k}) \crff \Psibardot_{j} -  \\&
         (\I_{k} \v_{k}) \crff \v_{k}\times \Phibar_{j}   +   \\& (\v_{k} \times^*) (\I_{k} \v_{k}) \crff \Phibar_{j} + \\&
         (\v_{k} \times^*) \I_{k}(\Psibardot_{j}) +  (\I_{k} \xibar_{k} ) \crff \Phibar_{j}  +  \\&
         \I_{k} \v_{\lambda(j)} \times \Psibardot_{j} -  \I_{k}( \v_{k} \times \Psibardot_{j})  + \\&
         \I_{k}\xibar_{\lambda(j)} \times \Phibar_{j}  \Big]  
        \label{partial_Cqdot_9}
      \end{aligned}
\end{equation}

Using the property P3 to combine terms,

   \begin{equation}
      \begin{aligned}
         \frac{\partial  [\,\C \qd\,]_{i}}{\partial \q_{j}} &=   \Phibar_{i}^{T}  \sum_{k \succeq j} \Big[  (\I_{k} \v_{k}) \crff \Psibardot_{j} +\\&
         (\v_{k} \times^* \I_{k} \v_{k} +\I_{k}  \xibar_{k}) \crff\Phibar_{j}  + \\&
         (\v_{k} \times^*) \I_{k}(\Psibardot_{j})  + \\&
         \I_{k} \v_{\lambda(j)} \times 
         \Psibardot_{j} - \\&
         \I_{k}( \v_{k} \times \Psibardot_{j})  + \I_{k}\xibar_{\lambda(j)} \times \Phibar_{j}  \Big]  
        \label{partial_Cqdot_10}
      \end{aligned}
\end{equation}
 
Summing over the index $k$, we get:

   \begin{equation}
      \begin{aligned}
         \frac{\partial  [\C \qd\,]_{i}}{\partial \q_{j}} &=     \Phibar_{i}^{T}  \Big[  2 \B_{j}^{C} \Psibardot_{j}  +     \I_{j}^{C} \v_{\lambda(j)} \times \Psibardot_{j}   + \\&
         \I_{j}^{C} \xibar_{\lambda(j)} \times \Phibar_{j}  + \zetabar_{j}^{C} \crff \Phibar_{j} \Big]  
        \label{partial_Cqdot_11}
      \end{aligned}
\end{equation}
 
\end{enumerate}

\vspace{1ex}
{\noindent \bf Partial Derivative of the Gravity Term:}

\begin{enumerate}
    \item Case when $j \preceq i$ \\
    For the case $j \preceq i$, the partial derivative of $\g_{i}$ with respect to $\q_{j}$ is:

\begin{equation}
     \frac{\partial \g_{i}}{ \partial \q_{j}} = \frac{\partial \Phibar_{i}^{T} }{\partial \q_{j}} \sum_{k \succeq i} \I_{k} \a_{0} + \Phibar_{i}^{T} \sum_{k \succeq i} \frac{\partial\I_{k} }{\partial \q_{j}} \a_{0}  
    \label{gFO_eqn_1}
\end{equation}
Using identities J3 and J4, we get:

\begin{equation}
    \begin{aligned}
   \frac{\partial \g_{i}}{ \partial \q_{j}} = -\Phibar_{i}^{T} \sum_{k \succeq i} (\I_{k} \a_{0}) \crff  \Phibar_{j} + \\
    \Phibar_{i}^{T} \sum_{k \succeq i} (\I_{k} \a_{0}) \crff \Phibar_{j} + \I_{k}(\a_{0} \times \Phibar_{j}) 
     \end{aligned}
    \label{gFO_eqn_2}
\end{equation}

Cancelling terms, and summing over index $k$, we get:

\begin{equation}
     \frac{\partial \g_{i}}{ \partial \q_{j}} = \Phibar_{i}^{T}  \I_{i}^{C}(\a_{0} \times \Phibar_{j})
    \label{gFO_eqn_3}
\end{equation}

    \item  Case when $j \succ i$ \\
    For the case $j \succ i$, we follow the similar process and use identity J4 as:

\begin{equation}
     \frac{\partial \g_{i}}{ \partial \q_{j}} =  \Phibar_{i}^{T} \sum_{k \succeq i} \frac{\partial\I_{k} }{\partial \q_{j}} \a_{0} 
    \label{gFO_eqn_4}
\end{equation}

Expanding: 

\begin{equation}
     \frac{\partial \g_{i}}{ \partial \q_{j}} =  \Phibar_{i}^{T} \sum_{k \succeq j} (\I_{k} \a_{0}) \crff \Phibar_{j} +\I_{k}(\a_{0} \times \Phibar_{j}) 
    \label{gFO_eqn_5}
\end{equation}

Summing over the index $k$, we get:

\begin{equation}
     \frac{\partial \g_{i}}{ \partial \q_{j}} = \Phibar_{i}^{T} \Big[(\I_{j}^{C} \a_{0})\crff \Phibar_{j}+ \I_{j}^{C}(\a_{0} \times \Phibar_{j}) \Big]
    \label{gFO_eqn_6}
\end{equation}
    
\end{enumerate}

\subsection{Details of Combining Terms (Partial Derivatives w.r.t $\q$):}
\label{combine_terms}

\begin{enumerate}
    \item $\frac{\partial \taubar_{i}}{\partial \q_{j}}$

Collecting the terms for $\frac{\partial [ \M \qdd]_{i}}{\partial \q_{j}}$, $  \frac{\partial  [\,\C \qd\,]_{i}}{\partial \q_{j}}$, and $ \frac{\partial \g_{i}}{ \partial \q_{j}}$ we get:

\begin{equation}
    \begin{aligned}
    \frac{\partial \taubar_{i}}{\partial \q_{j}} &= \Phibar_{i}^{T} [  \I_{i}^{C}  (\gammabar_{\lambda(j)} )  \times \Phibar_{j} ]  + \\&~~~~
    \Phibar_{i}^{T}  \Big[  2 \B_{i}^{C} \Psibardot_{j}  +     \I_{i}^{C} \v_{\lambda(j)} \times \Psibardot_{j}   + \\&~~~~
    \I_{i}^{C} \xibar_{\lambda(j)} \times \Phibar_{j}  \Big]  + 
   \Phibar_{i}^{T}  \I_{i}^{C}(\a_{0} \times \Phibar_{j})
    \end{aligned}
\end{equation}

Re-arranging terms here

\begin{equation}
    \begin{aligned}
        \frac{\partial \taubar_{i}}{\partial \q_{j}} = \Phibar_{i}^{T} [  \I_{i}^{C}  (\xibar_{\lambda(j)}+\gammabar_{\lambda(j)}+\a_{0} ) \times \Phibar_{j}   +\\
        2 \B_{i}^{C} \Psibardot_{j}  +  \I_{i}^{C} \v_{\lambda(j)} \times \Psibardot_{j}]
        \end{aligned}
\end{equation}

Simplifying:

\begin{equation}
    \begin{aligned}
        \frac{\partial \taubar_{i}}{\partial \q_{j}} = \Phibar_{i}^{T} [  \I_{i}^{C}  \a_{\lambda(j)} \times \Phibar_{j}   +   2 \B_{i}^{C} \Psibardot_{j}  +  \\
        \I_{i}^{C} \v_{\lambda(j)} \times \Psibardot_{j}]     
    \end{aligned}
\end{equation}

\begin{equation}
    \begin{aligned}
    \frac{\partial \taubar_{i}}{\partial \q_{j}} = \Phibar_{i}^{T} [  \I_{i}^{C}  \a_{\lambda(j)}   \times \Phibar_{j}  +  2 \B_{i}^{C} \Psibardot_{j}  + \\ \I_{i}^{C} \v_{\lambda(j)} \times \Psibardot_{j}]     
    \end{aligned}
\end{equation}

Using $\Psibarddot_{k}$ (Eq.~\ref{psidot_psidotdot_defn}), the final compact expression for $\frac{\partial \taubar_{i}}{\partial \q_{j}} $ is:

\begin{equation}
    \frac{\partial \taubar_{i}}{\partial \q_{j}} =  \Phibar_{i}^{T} \big[ 2  \B_{i}^{C} \big] \Psibardot{}_{j} + \Phibar_{i}^{T}  \I_{i}^{C} \Psibarddot{}_{j}
    \label{dtau_dq_eqn1}
\end{equation}

\item $\frac{\partial \taubar_{j}}{\partial \q_{i}}  (j \neq i)$

\vspace{1ex}
Similar to the previous case, collecting the terms  $\frac{\partial [ \M \qdd]_{j}}{\partial \q_{i}}$, $  \frac{\partial  [\,\C \qd\,]_{i}}{\partial \q_{i}}$, and $ \frac{\partial \g_{j}}{ \partial \q_{i}}$ we get:

\begin{equation}
 \begin{aligned}
    \frac{\partial \taubar_{j}}{\partial \q_{i}} = \Phibar_{j}^{T} [\etabar_{i}^{C} \crff \Phibar_{i} + \I_{i}^{C} (\gammabar_{\lambda(i)})\times \Phibar_{i}  ] + \\
    \Phibar_{j}^{T}  \Big[  2 \B_{i}^{C} \Psibardot_{i}  +  \I_{i}^{C} \v_{\lambda(i)} \times \Psibardot_{i} +  \\
    \I_{i}^{C} \xibar_{\lambda(i)} \times \Phibar_{i}  + 
    \zetabar_{i}^{C} \crff \Phibar_{i} \Big]  + \\
    \Phibar_{j}^{T} [(\I_{i}^{C}\a_{0})\crff \Phibar_{i}+  \I_{i}^{C}(\a_{0} \times \Phibar_{i}) ]
 \end{aligned}
\end{equation}

Re-arranging terms, we get:

\begin{equation}
 \begin{aligned}
    \frac{\partial \taubar_{j}}{\partial \q_{i}} = \Phibar_{j}^{T} [ (\etabar_{i}^{C} +\zetabar_{i}^{C} +\I_{i}^{C}\a_{0})\crff \Phibar_{i} + \\
    \I_{i}^{C} (\gammabar_{\lambda(i)}  +\xibar_{\lambda(i)}+\a_{0})\times \Phibar_{i}  + \\
    2 \B_{i}^{C} \Psibardot_{i}  + \I_{i}^{C} \v_{\lambda(i)} \times \Psibardot_{i}  ]
 \end{aligned}
\end{equation}

Using the definition of  $\a_{i}$ and $\f_{i}$ (Eq.~\ref{f_defn}), we get:

\begin{equation}
    \begin{aligned}
        \frac{\partial \taubar_{j}}{\partial \q_{i}} = \Phibar_{j}^{T} [ (\f_{i}^{C} )\crff \Phibar_{i} + \I_{i}^{C} \a_{\lambda(i)}\times \Phibar_{i}  + \\
        2 \B_{i}^{C} \Psibardot_{i}  +   \I_{i}^{C} \v_{\lambda(i)} \times \Psibardot_{i}  ]
    \end{aligned}
\end{equation}

Using the definition of $\Psibarddot_{i}$ (Eq.~\ref{psidot_psidotdot_defn}), we get the final expression for $ \frac{\partial \taubar_{j}}{\partial \q_{i}}$ as:

\begin{equation}
    \frac{\partial \taubar_{j}}{\partial \q_{i}} = \Phibar_{j}^{T} [ 2 \B_{i}^{C} \Psibardot_{i} +  \I_{i}^{C}  \Psibarddot_{i}+(\f_{i}^{C} )\crff \Phibar_{i} ]
    \label{dtau_dq_eqn2}
\end{equation}

\end{enumerate}

\subsection{Partial Derivatives of $ID$ w.r.t $\qd$: Derivations}
\label{partials_qd}

\begin{enumerate}
    \item Case when $j \preceq i$ \\
    
      Using product rule of differentiation in Eq.~\ref{cqdot_exp_2}, we get:
  \begin{equation}
      \begin{aligned}
        \frac{\partial  [\,\C \qd\,]_{i}}{\partial \qd_{j}} =   \Phibar_{i}^{T}  \sum_{k \succeq i} \Bigg[\frac{\partial (\v_{k} \times^*)}{\partial \qd_{j}} \I_{k} \v_{k} + \\
         \v_{k} \times^* \I_{k} \frac{ \partial (\v_{k})}{\partial \qd_{j}} +  \I_{k} \frac{\partial \xibar_{k} }{\partial \qd_{j}} \Bigg]  
        \label{partial_Cqdot_dot_1}
      \end{aligned}
\end{equation}

Using identities J8 and J9, we get:

\begin{equation}
      \begin{aligned}
        \frac{\partial  [\,\C \qd\,]_{i}}{\partial \qd_{j}} =   \Phibar_{i}^{T}  \sum_{k \succeq i} \Big[(\I_{k} \v_{k}) \crff \Phibar_{j} 
        + \v_{k} \times^* \I_{k} \Phibar_{j} + \\
        \I_{k} ( (\v_{\lambda(j)} - \v_{k})  \times \Phibar_{j} + \Phibardot_{j} ) \Big]  
        \label{partial_Cqdot_dot_2}
      \end{aligned}
\end{equation}

Simplifying and collecting terms, we get:

  \begin{equation}
      \begin{aligned}
        \frac{\partial  [\,\C \qd\,]_{i}}{\partial \qd_{j}} =   \Phibar_{i}^{T}  \sum_{k \succeq i} \Big[(\I_{k} \v_{k}) \crff \Phibar_{j}  + \\ \v_{k} \times^* \I_{k} \Phibar_{j}  -
        \I_{k}(\v_{k} \times) \Phibar_{j} +\\
        \I_{k} (\v_{\lambda(j)}  \times \Phibar_{j} +  \Phibardot_{j} ) \Big]  
        \label{partial_Cqdot_dot_3}
      \end{aligned}
\end{equation}

Summing over the index $k$, we get:

  \begin{equation}
      \begin{aligned}
        \frac{\partial  [\,\C \qd\,]_{i}}{\partial \qd_{j}} =   \Phibar_{i}^{T} \Big[2 \B_{i}^{C} \Phibar_{j} +  \I_{i}^{C} (\Psibardot_{j} + \Phibardot_{j} ) \Big]  
        \label{partial_Cqdot_dot_4}
      \end{aligned}
\end{equation}

\item Case when $j \succ i$ 

  Similar to the previous case, product rule of differentiation in Eq.~\ref{cqdot_exp_2} is used:
  
  \begin{equation}
      \begin{aligned}
        \frac{\partial  [\,\C \qd\,]_{i}}{\partial \qd_{j}} =   \Phibar_{i}^{T}  \sum_{k \succeq i} \Big[\frac{\partial (\v_{k} \times^*)}{\partial \qd_{j}} \I_{k} \v_{k} + \\
         \v_{k} \times^* \I_{k} \frac{ \partial (\v_{k})}{\partial \qd_{j}} +  \I_{k} \frac{\partial \xibar_{k} }{\partial \qd_{j}} \Big]  
        \label{partial_Cqdot_dot_5}
      \end{aligned}
\end{equation}

Using identities J8 and J9, we get:

  \begin{equation}
       \begin{aligned}
        \frac{\partial  [\,\C \qd\,]_{i}}{\partial \qd_{j}} &=   \Phibar_{i}^{T}  \sum_{k \succeq j} \Bigg[(\I_{k} \v_{k}) \crff \Phibar_{j}  + \v_{k} \times^* \I_{k} \Phibar_{j} + \\&
         \I_{k} ((\v_{\lambda(j)} - \v_{k})  \times \Phibar_{j} + 
        \Phibardot_{j} ) \Bigg]  
        \label{partial_Cqdot_dot_6}
      \end{aligned}
\end{equation}

Simplifying and collecting terms, we get:

   \begin{equation}
      \begin{aligned}
        \frac{\partial  [\,\C \qd\,]_{i}}{\partial \qd_{j}} =   \Phibar_{i}^{T}  \sum_{k \succeq j} \Big[(\I_{k} \v_{k}) \crff \Phibar_{j}  + \\
        \v_{k} \times^* \I_{k} \Phibar_{j}  -\I_{k}(\v_{k} \times) \Phibar_{j} + \\
        \I_{k} (\v_{\lambda(j)}  \times \Phibar_{j} + \Phibardot_{j} ) \Big] 
        \label{partial_Cqdot_dot_7}
      \end{aligned}
\end{equation}

Summing over the index $k$, we get:

  \begin{equation}
      \begin{aligned}
        \frac{\partial  [\,\C \qd\,]_{i}}{\partial \qd_{j}} =   \Phibar_{i}^{T} \Big[2 \B_{j}^{C} \Phibar_{j} +  \I_{j}^{C} (    \Psibardot_{j} + \Phibardot_{j} ) \Big]  
        \label{partial_Cqdot_dot_8}
      \end{aligned}
\end{equation}
\end{enumerate}

Flipping the indices $i$ and $j$ in Eq.~\ref{partial_Cqdot_dot_8} to get $\frac{\partial  [\,\C \qd\,]_{j}}{\partial \qd_{i}} $ for the case $j \prec i$:
  
  \begin{equation}
      \begin{aligned}
        \frac{\partial  [\,\C \qd\,]_{j}}{\partial \qd_{i}} =   \Phibar_{j}^{T} \Big[2 \B_{i}^{C} \Phibar_{i} +  \I_{i}^{C} (    \Psibardot_{i} + \Phibardot_{i} ) \Big]  (j \neq i)
        \label{partial_Cqdot_dot_9}
      \end{aligned}
\end{equation}

Hence, the partial derivatives of $\taubar$ with respect to $\qd$ are:

  \begin{equation}
      \begin{aligned}
      &  \frac{\partial  \taubar_{i}}{\partial \qd_{j}} =   \Phibar_{i}^{T} \Big[2 \B_{i}^{C} \Phibar_{j} +  \I_{i}^{C} (    \Psibardot_{j} + \Phibardot_{j} ) \Big]  \\
      &   \frac{\partial  \taubar_{j}}{\partial \qd_{i}} =   \Phibar_{j}^{T} \Big[2 \B_{i}^{C} \Phibar_{i} +  \I_{i}^{C} (    \Psibardot_{i} + \Phibardot_{i} ) \Big]  (j \neq i)
        \label{tau_FO_qd_eqn1}
      \end{aligned}
\end{equation}

}

\end{document}